 \documentclass[sigconf]{acmart}

\AtBeginDocument{%
  \providecommand\BibTeX{{%
    \normalfont B\kern-0.5em{\scshape i\kern-0.25em b}\kern-0.8em\TeX}}}

\setcopyright{acmlicensed}
\copyrightyear{2025}
\acmYear{2025}
\acmDOI{XXXXXXX.XXXXXXX}

\acmConference[CHI '25]{Proceedings of the 2025 CHI Conference on Human Factors in Computing Systems}{April 26-- May 1, 2025}{Yokohama, Japan}
\acmISBN{978-1-4503-XXXX-X/18/06}
\usepackage{float}
\usepackage{enumitem}
\usepackage{multirow} 
\usepackage{graphicx} 
\usepackage{array}    

\newcommand{\revision}[1]{\textcolor{black}{#1}}

\usepackage{framed}

\copyrightyear{2025} 
\acmYear{2025} 
\setcopyright{cc}
\setcctype{by}
\acmConference[CHI '25]{CHI Conference on Human Factors in Computing Systems}{April 26-May 1, 2025}{Yokohama, Japan}
\acmBooktitle{CHI Conference on Human Factors in Computing Systems (CHI '25), April 26-May 1, 2025, Yokohama, Japan}\acmDOI{10.1145/3706598.3713201}
\acmISBN{979-8-4007-1394-1/25/04}

\begin{document}

\title[Choice Independence in Using Multilingual LLMs for Persuasive Co-Writing Tasks in Different Languages]{Mind the Gap! Choice Independence in Using Multilingual LLMs for Persuasive Co-Writing Tasks in Different Languages}



\author{Shreyan Biswas}
\affiliation{%
  \institution{Delft University of Technology}
  \city{Delft}
  \country{The Netherlands}}
\email{s.biswas@tudelft.org}

\author{Erlei, Alexander}
\affiliation{%
  \institution{University of Goettingen}
  \city{Goettingen}
  \country{Germany}}
\email{alexander.erlei@wiwi.uni-goettingen.de}

\author{Ujwal Gadiraju}
\affiliation{%
  \institution{Delft University of Technology}
  \city{Delft}
  \country{The Netherlands}}
\email{u.k.gadiraju@tudelft.nl}

\renewcommand{\shortauthors}{Biswas et al.}

\begin{abstract}
Recent advances in generative AI have precipitated a proliferation of novel writing assistants. These systems typically rely on multilingual large language models (LLMs), providing globalized workers the ability to revise or create diverse forms of content in different languages. However, there is substantial evidence indicating that the performance of multilingual LLMs varies between languages. Users who employ writing assistance for multiple languages are therefore susceptible to disparate output quality. Importantly, recent research has shown that people tend to generalize algorithmic errors across independent tasks, violating the behavioral axiom of \textit{choice independence}. In this paper, we analyze whether user utilization of novel writing assistants in a charity advertisement writing task is affected by the AI's performance in a second language. Furthermore, we quantify the extent to which these patterns translate into the persuasiveness of generated charity advertisements, as well as the role of peoples' beliefs about LLM utilization in their donation choices. Our results provide evidence that writers who engage with an LLM-based writing assistant violate choice independence, as prior exposure to a Spanish LLM reduces subsequent utilization of an English LLM. While these patterns do not affect the aggregate persuasiveness of the generated advertisements, people's beliefs about the source of an advertisement (human versus AI) do. In particular, Spanish-speaking female participants who believed that they read an AI-generated advertisement strongly adjusted their donation behavior downwards. Furthermore, people are generally not able to adequately differentiate between human-generated and LLM-generated ads. Our work has important implications for the design, development, integration, and adoption of multilingual LLMs as assistive agents---particularly in writing tasks.

\end{abstract}

\ccsdesc[500]{Human-centered computing~Empirical studies in HCI}
\ccsdesc[300]{Human-centered computing~User studies}
\ccsdesc[300]{Applied computing~Economics}
\ccsdesc[100]{Applied computing~Psychology}
\ccsdesc[300]{Human-centered computing~Empirical studies in collaborative and social computing}
\keywords{Human-AI interaction, Choice Independence, Multilingual LLMs, User Reliance}

\maketitle
\section{Introduction}

\revision{The advent of large language models (LLMs) has transformed human-AI co-writing \cite{Reza2023-hp}. Modern writing assistants such as Microsoft 365 Copilot, Grammarly, or Jasper leverage multilingual LLMs to support a global user base in drafting, editing, and rewriting content \cite{noy2023experimental,herbold2023large}. However, despite their widespread use, LLMs still differ substantially between languages \cite{Ahuja2023-zu,Zhang2024-mp,huang2023not,bang2023multitask,joshi2020state,jiao2023chatgpt}. While English is the default language and therefore exhibits robust performance, lower resource languages often display deficiencies in fluency, coherence, sensitivity to jailbreaks, and contextual appropriateness \cite{conde2024opensourceconversationalllms, yong2023low, hasan2024largelanguagemodelsspeak, lawalcontextual}.}

\revision{These performance disparities in multilingual LLMs can have significant \textbf{behavioral implications}. Users interacting with LLMs in multilingual settings may not only be influenced by technical limitations but also by their perceptions and experiences from being exposed to heterogeneous output quality. Consider a bilingual journalist writing news articles in both English and Spanish. When using an LLM assistant, they notice it struggles with Spanish idioms and cultural references, requiring substantial editing. A rational writer would evaluate each language independently, maximizing the LLM's benefits in English while being more selective in Spanish. Yet, recent evidence has shown that humans use prior experience with LLMs to predict future performance \cite{rambachan2024llms} and, in particular, often violate the independence axiom of rational choice theory by generalizing AI errors to objectively unrelated tasks \cite{erlei2024understanding}. In the example above, this may cause the journalist to avoid the LLM for English articles due to a faulty generalization of their Spanish experience. Similar arguments apply to all regions or domains where linguistic diversity is high, such as customer support, technical documentation, or professional writing. Understanding how users react to heterogeneity in these scenarios is important, particularly given the current trend where LLM assistants are incrementally rolled out in multiple languages to meet global demands \cite{bensum2024languages,meta2024ai,venkatachary2024ai}.}

\revision{Importantly, there are good reasons to doubt that these behavioral patterns would automatically resolve solely through experience and learning-by-doing. First, human-AI interaction research highlights the critical role of \textbf{first impressions} and initial exposures for user behaviour and trust \cite{schwartz2023enhancing,nourani2020investigating,glikson2020human,tolmeijer2021second}. Even with repeated zero-stakes interactions, individuals frequently fail to optimally calibrate their reliance on AI systems after observing initial errors, exhibiting persistent downward bias in utilization patterns. In real-world settings, learning about LLM capabilities is costly, which can further deter users from updating their beliefs. Second, it is unclear whether choice independence violations stem from biased information updating or a biased choice rule. While the latter may improve over time, biased information updating inhibits learning and hence perpetuates inefficient under-utilization.}

In our study,\footnote{The codebase and datasets are openly available: https://sites.google.com/view/cillm-mind-the-gap} we examine how exposure to two different languages affects people's interactions with an LLM-based writing assistant in the context of persuasive writing. We focus on English and Spanish, two high-resource languages that allow us to commission multiple advertisements from writers and subsequently test them in a charitable giving task. Participants in the first experiment wrote two advertisements for the charity World Wildlife Fund (WWF), one in English and one in Spanish, with the explicitly incentivized assignment to make it as persuasive as possible. By varying the order of the language, we measure how exposure to the lower (higher) resource language Spanish (English) affects subsequent interaction with our LLM-based co-writing tool in the higher (lower) resource language English (Spanish). For comparison, we also ask native English and Spanish speakers to write advertisements without any AI assistance. The second experiment then evaluates the generated ads via a charitable giving task in which participants split an endowment of \pounds1.5 between themselves and the WWF after reading a treatment-dependent advertisement. In addition to the ads from the first experiment, we also evaluate advertisements that are purely derived from LLMs, as well as the WWF mission statement. Finally, we elicit donors' beliefs about the origin of the advertisement (human versus AI). Through this empirical setup, 
we address the following four research questions:

\begin{framed}
\begin{itemize}
    \item [\textbf{RQ 1:}] How does LLM performance in one language affect user utilization in a second language for persuasive co-writing?

    \item [\textbf{RQ 2:}] How do varying levels of LLM utilization in co-writing tasks across languages influence the persuasiveness of generated advertisements?

    \item [\textbf{RQ 3:}]
    How does altruism persuasiveness differ between human writers, human-LLM teams, and LLMs?

    \item [\textbf{RQ 4:}] How do donor beliefs about the source of an advertisement affect altruistic behaviour?
    
\end{itemize}
\end{framed}

Our results confirm that writers violate choice independence by generalizing their experience with one language to a second language and adapting their behaviour accordingly. In particular, being first exposed to the lower-resource Spanish AI assistant reduces subsequent reliance on the English LLM. There is also moderate evidence that writers who first experience the English LLM may use the Spanish writing assistant more in a follow-up task. Experiment 2 shows that these differences do not affect the persuasiveness of the generated advertisements. We find that donations are very stable across conditions, suggesting that participants' altruistic donation preferences are largely independent of the kind of advertisement we utilize in this study. However, we found moderate evidence that sole human writers may be less effective in eliciting donations, which is absent from human-LLM teams. Finally, participants cannot reliably identify whether an advertisement was generated by a human or an LLM, but may still condition their donation behavior on these beliefs. Here, we document strong cultural and gender effects, as female Spanish-speaking participants who believe that they are reading an AI-generated ad (1) donate substantially less and (2) are much more likely to not donate at all. For all other groups, results qualitatively point in the same direction, but are much smaller and not statistically significant. In combination with the fact that female participants are significantly more likely to donate, and donate more, human reactions to persuasive text that is perceived to be generated by LLMs are likely highly context-dependent.

In general, our results have strong implications for the understanding, design and deployment of multilingual AI assistants. Following prior abstract work on user violations of choice independence, this is the first study replicating these results in an applied context, highlighting the need for theory to consider how humans systematically deviate from rational choice theory when exposed to AI output with heterogeneous quality across tasks. In particular, users do not appear to evaluate LLMs tasks independently, but holistically, even though the latter approach leads to faulty perceptions. Second, for practitioners, it is important to consider potential unintended second-order effects when deploying multilingual systems that vary significantly in quality. Those who utilize features in more than one language may adjust utilization downwards, decreasing demand. Employers themselves should also be mindful of how these patterns might shape employee performance, and adjust accordingly. From a societal perspective, there are two immediate implications. One, the adoption of LLMs may lag in non-English-speaking countries (particularly for those with low-resource native languages), and two, choice independence violations may increase inequity between countries with high- and low-resource languages. As high-resource languages are usually positively related to the native country's economic position, this implies potentially lower uptake in less wealthy countries, or from immigrated employees within wealthy countries, plausibly increasing inequality \cite{noy2023experimental}. Here, it is important to note that our results should provide a lower baseline for real-world impact, as Spanish is generally considered one of the highest resource languages. \revision{While it is possible that starker differences also make it easier for users to categorically distinguish between languages and thereby increase rational behavior, prior work suggests that users violate choice independence in the context of AI even after observing two distinct error graphs for two distinct problems \cite{erlei2024understanding}. This suggests that differentiation alone does not alleviate biased generalizations.} Finally, our results about donation behaviour as a response to beliefs about LLM-generated advertisements echo recent work whereby humans tend to prefer engagement with other humans and react adversely to AI or algorithm-generated output \cite{zhang2023human,bellaiche2023humans,millet2023defending,erlei2022s}, \revision{while also being unable to adequately identify it \cite{kobis2021artificial}.} Practitioners aiming to elicit donations may benefit strongly from \revision{reducing uncertainty around the use of AI in the context of charitable giving, e.g., through promoting the salience of humans in the context of persuasion or advertising.}

\section{Background and Related Work}
Our study integrates concepts from multilingual large language models, charitable giving, (persuasive) human-AI co-writing, expected utility theory, and human perceptions of AI-generated content. We present related literature in these realms and we position our work and contributions at their confluence.

\subsection{Multilingual LLMs}
Large Language Models have demonstrated strong multilingual capabilities across numerous tasks \cite{zhao2023survey,bang2023multitask,le2023bloom}, which has led to their widespread proliferation across various countries \cite{Kaddour_2023}. Many multinational companies rely on their translations to accelerate cross-national team cooperation, and individual workers across the globe benefit from individualized LLM assistance that enhances their productivity -- both in their native and the English language. Yet, despite their immense promise, LLMs still differ substantially across languages \cite{Ahuja2023-zu,Zhang2024-mp,huang2023not,bang2023multitask,joshi2020state,jiao2023chatgpt,hada2024akal}. LLMs not only show poorer text generation and problem-solving performance for low-resource languages, but also heightened security vulnerabilities, safety challenges, and tokenizer biases \cite{Shen_2024,ahia2023languagescostsametokenization,Ahuja2023-zu}.

These caveats are not strictly limited to languages conventionally considered low-resource. Multiple open-source LLMs are disproportionately trained on English data. For example, PaLM2's training corpus consists of 78.99\% English data, compared to only 2.11\% for Spanish data \cite{chowdhery2023palm}, and LLAMA-2 exhibits a mere 0.11\% Spanish representation \cite{touvron2023llama}. \revision{Despite these gaps, Spanish is generally considered a relatively high-resource language. The GPT-4 technical report \cite{achiam2023gpt} shows benchmark accuracies of 85.5\% and 84\% for English and Spanish respectively in the multilingual version of the MMLU \cite{hendrycks2021ethics}. Spanish also performs relatively well in the QWEN2 technical report for professional annotator tasks \cite{yang2024qwen2}.}

\revision{The practical performance of multilingual LLMs in the Spanish language, however, is often relatively poor, especially in contextual usage and practical applications \cite{jin2024better, conde2024opensourceconversationalllms, zhang2024dolares}. A particularly striking finding is highlighted by \citet{conde2024opensourceconversationalllms}, who show that most open-source LLMs exhibit significant comprehension deficiencies for the Spanish vocabulary. Two-thirds of the models, including the Llama-2 series (a predecessor of the Llama 3.1 model used in our experiments), fail to provide valid definitions for more than 50\% of tested Spanish words. Moreover, when evaluated for contextual word usage, most models fall below 10\%. For instance, the Llama-2-7b model correctly defines only 42 out of 100 words and uses just 3 out of 100 words correctly in context. Importantly, these failures are not confined to low-frequency words; even highly frequent words like "minuto" fail in meaning across 8 out of 12 evaluated models. While their work highlights linguistic limitations, our research extends this by exploring the behavioural and real-world impacts of such deficiencies.}
Particularly in co-creation environments that span linguistic and cultural boundaries, varying performance levels can disrupt the collaborative process, leading to asymmetric misunderstandings of system capabilities and thereby in inappropriate reliance that hurts efficiency \cite{Dhillon2024-ol}. This may be particularly problematic in sensitive domains such as persuasive writing, which depend on subtle combinations of tangible and non-tangible elements, like emotional appeals and accurate fact specificities \cite{Hibbert2007-ty,Choi2020-eh,Wymer2023-pg,Dafouz-Milne2008-iy}. 
\revision{However, the ground reality is that multilingual LLMs are largely being deployed and used immaterial of their performance in specific languages. And while companies often do evaluate their models across languages, traditional technical benchmarks may not capture the full picture due to downstream consequences for user behaviour -- even when performance across languages appears comparable. Inspired by this real-world context, we explore the utilization of multilingual LLMs in a co-writing task and study how exposure effects with LLMs in different languages influence user interaction and behavior.}

\subsection{Charitable Persuasive Writing}

Persuasive writing is a form of communication that seeks to convince the reader to adopt a particular viewpoint or take a specific action \cite{jonsen2018convincing}. It can be viewed through the lens of sender-receiver games, a concept generated from economic theory frequently applied in computer science across domains such as recommender systems \cite{Apel2020-tt}, reinforcement learning \cite{hiraoka2014reinforcement}, multi-agent interaction \cite{meta2022human}, and most recently LLMs \cite{Shin_Kim_2024}.

Advertisers often leverage this sender-receiver model (e.g., by employing relevant recommender systems) to influence a receiver’s behaviour by appealing to emotions or raising awareness, and charitable advertisement writing is no exception \cite{salvi2024conversational, furumai2024zero,Murphy2001-yn}.

With the rise of LLMs, the traditional dynamic of a human agent influencing a human receiver has evolved. Now, LLM agents can serve as persuasive entities, sometimes more efficiently and effectively than humans \cite{Breum2024-xf, Matz2024-pl, Zeng2024-lp, Xu2023-bs, Goldstein2023-wr,durmus2024measuring}. LLM-generated text has been shown to influence political attitudes \cite{voelkel2023artificial}, vaccine uptake \cite{karinshak2023working}, strategic negotiations \cite{meta2022human}, personal beliefs \cite{salvi2024conversational}, or even romantic conversations \cite{zhou2020design}. However, LLMs also exhibit many limitations -- such as hallucination, a lack of contextual knowledge and wordiness \cite{Myers_2024}--- prompting a shift towards co-creation where human and AI agents collaborate as persuasive agents to ensure both effectiveness and reliability \cite{Dhillon2024-ol, zhang2023human}. Specifically in the context of altruistic social preferences such as charitable giving, little is known about the effectiveness of LLMs in eliciting donations. Even the psychological literature is fragmented, lacking a coherent model about which factors specifically increase the effectiveness of charity advertisement \cite{xu2020relative,saeri2023works,schamp2023effectiveness,fan2020factors}. This makes charitable giving not only a novel but potentially high-value field of application for persuasive LLMs. We, therefore, consider the task of charitable persuasive writing as a lens to study user behavior with multilingual LLMs in this paper.

\subsection{LLM Augmented Co-writing}
There has been considerable interest from the HCI community in analyzing and fostering collaborative human-AI writing. Many popular real-world applications like Microsoft Word and Gmail already utilize smart features that, e.g., predict the next words a user is likely to write or provide context-dependent phrasing advice (auto-completion suggestions). LLMs themselves have demonstrated exceptional performance in open-ended writing, with great potential benefit for a wide variety of tasks \cite{lee2022coauthor,macneil2022automatically,meyer2022we,mirowski2023co,yuan2022wordcraft,zhao2023more,herbold2023ai}.

Beyond evaluating the output of foundational LLMs, the HCI community has begun to create tools designed to support human writers. \citet{Kim2023-wn} introduce a framework that augments ``object-oriented'' interaction with LLMs. Their approach enables end-users to ``track, compare, and combine'' configurations, fostering inspiration and creativity in the design process. Based on this framework, there has been an ongoing development of novel and innovative systems, focusing on integral parts of the writing process such as non-linearity, rewriting or idea-generation \cite{Reza2023-hp,lu2024corporate, teufelberger2024llm}.

In this paper, we utilize and slightly modify ABScribe, a tool created by \citet{Reza2023-hp}, which allows users to ``track, compare, and modify variations" while interacting directly with an LLM system to generate new ideas during the writing process. The system is purposefully designed to allow for a seamless writing flow by providing users with flexibility and autonomy throughout the co-creating process. Through different functions that go beyond mere text generation, it provides users with a more targeted approach to exploit the various use cases of LLMs. 

The integration of LLMs into the writing process has the potential to transform how content is created across various fields. However, the effectiveness of LLM-augmented co-writing depends on several factors, including the quality of the LLM, the nature of the task, the expertise of the human writer, the interaction of the human writer with the LLM, and the perceptions of consumers towards LLMs. In this paper, we focus on the latter two aspects, as both appropriate reliance and consumer attitudes have been previously shown to be important facets in human-AI interaction  \cite{He_2023,sara_23,schemmer2023appropriate,erlei2024understanding,wester2024exploring,grassini2024understanding}.

\subsection{Choice Independence and Human-AI Interaction}
Our study focuses on choice independence as a particularly important aspect of human-AI interaction.\footnote{The von Neumann-Morgenstern's utility theory (or expected utility theory), provides a key mechanism for understanding the behaviour of a rational agent under uncertainty. In this framework, a rational agent makes decisions that maximize the subjective value of their utility when faced with stochastic outcomes \cite{vnm-original}. The theory is built upon four main axioms: \textit{completeness}, \textit{transitivity}, \textit{independence}, and \textit{continuity}. Among these, the independence axiom is the most contentious and has significant implications for rational decision-making \cite{Holt1986-jp}. 
Mathematically, this axiom is represented as follows,
\[
X \succ Y \implies pX + (1-p)Z \succ pY + (1-p)Z \quad \text{for} \quad 0 < p \leq 1
\]

Here, $X$, $Y$, and $Z$ represent lotteries, which can be thought of as stochastic processes or uncertain events that yield probability distributions over a set of outcomes. If amongst the lotteries $X$ and $Y$ a rational agent is said to prefer
lottery $X$ over $Y$ their preference should remain unchanged even if an irrelevant lottery is introduced and mixed with both $X$ and $Y$ in equal proportion. This axiom asserts the stability of preferences and is considered a cornerstone of rationality in decision theory.} 
In the context of multilingual LLMs, it postulates that users who experience two or more languages should evaluate them independently, adjusting their usage according to their language-specific experiences. Choice independence is a foundational axiom of expected utility theory \cite{vnm-original}, and often implicitly assumed when deploying novel technologies, systems and products \cite{erlei2024understanding,Ethayarajh2022-gj}. We argue that this is one of the reasons why companies tend to simultaneously deploy their AI assistants globally, despite the documented differences in performance across languages. The HCI literature has only recently begun to empirically scrutinize its applicability in the context of algorithmic and AI systems. \cite{pareek2024trust,erlei2024understanding}.

\revision{So far, evidence from HCI work is constrained to a limited number of abstract 
decision tasks. For example, \citet{erlei2024understanding} uses an online experiment to explore how humans delegate decisions to a superior AI system across two independent abstract prediction tasks. They manipulate the performance of the AI system such that participants either observe an AI system that provides the best-possible prediction in both tasks or one that exhibits a systematic error in one of the two. Results show that the induced error in one task significantly reduces trust in and delegation to the AI system even for the second prediction, indicating that participants erroneously generalize AI errors across tasks, violating choice independence and undermining appropriate reliance. Similarly, \citet{pareek2024trust} conducted an online experiment to study trust dynamics in the context of complementary Human-AI expertise. Participants engage in classification tasks involving familiar (High Human Expertise, HHE) and unfamiliar (Low Human Expertise, LHE) stimuli. The paper shows that people calibrate trust in the AI for LHE tasks based on its performance in HHE tasks, demonstrating a spillover of trust judgments across tasks. While these studies are informative, abstract decision tasks are not only specifically designed to artificially test a specific hypothesis while controlling for the entire context (e.g., expertise, task), but also benefit from very focused attention towards the specific problem a researcher is interested in. Real-world decisions are often more complicated, limiting the extent to which certain laboratory results can be generalized or scaled \cite{list2022voltage,brandon2022human,sara_23,salimzadeh2024dealing}. For example, in \citet{erlei2024understanding}, AI errors are precisely quantifiable and codified, effectively alleviating any participant uncertainty about model performance, and facilitating easy comparisons of the AI's performance across tasks. In real-world settings, many people learn by updating their beliefs solely through experience and noisy feedback, while always being uncertain about the model's ``true'' performance.
Therefore, in this paper, we extend the analysis of human behaviour in the context of heterogeneous AI output across distinct tasks to the applied context of human-AI co-writing}. Multilingual LLMs represent a prime example to test whether humans tend to rationally learn about and evaluate LLMs, because (1) multilingual writing is everywhere, (2) writing and text generation belong to the most common use-cases of LLMs, (3) writing is a complex and non-linear activity for which humans possess intimate familiarity and expertise, and (4) producing persuasive text in one language is distinct from the problem of producing persuasive text in a second language.

\subsection{Human Attitudes Towards AI Generated Content} There has been a lot of 
interdisciplinary literature that has analyzed how humans react to AI-generated output while articulating the notions of algorithmic affinity and aversion~\cite{dietvorst2015algorithm,dietvorst2018overcoming}. Within the scope of this paper, we are primarily interested in textual or persuasive content. In addition, eliciting donations through advertisements is closely related to negotiation scenarios. Recent studies provide mixed results on human perceptions of AI-generated content. In \citet{lim2024effect}, disclosing AI as the source of communication negatively impacts human perceptions of messages. People may prefer AI advertisements depending on which kind of appeal is made \cite{chen2024consumer}, but can react negatively towards AI use by charities \cite{arango2023consumer} and generally appear to denigrate creators who transparently use AI \cite{rae2024effects,bruns2024you}. Other research finds positive effects of revealing the use of AI technology in the context of influencing and persuasion \cite{wang2024positive} and no creator loss in credibility \cite{huschens2023you}. In general, the literature documents several divergent effects, currently lacking a parsimonious explanation \cite{ferraro2024paradoxes}. Interestingly, people often appear unable to identify AI-generated content \cite{clark2021all} and only reveal preferences against it upon disclosure \cite{kobis2021artificial}, possibly due to inherent pro-human attitudes \cite{grassini2024understanding,zhang2023human}. In bargaining and negotiations, humans also tend to exhibit preferences for other humans \cite{erlei2022s}, and behave more self-interested \cite{erlei2020impact,shen2024bargaining,chugunova2022we,von2023social}. We add to this existing bed of literature by examining how peoples' beliefs about the origin of a charitable advertisement affect donation behaviour, and whether these beliefs correlate with true LLM usage.

\section{System Design}
We used the ABScribe tool built by \citet{Reza2023-hp} as the foundational system for our study. As mentioned earlier ABScribe is designed to support object-oriented interaction \cite{Kim2023-wn} 
within an LLM-powered writing environment, enabling users to efficiently explore and organize multiple text variations while co-writing with LLMs.
The system provides two primary features that are crucial for our study:
\begin{enumerate}
    \item \textbf{AI Modifiers}: This feature allows users to modify their text based on predefined prompts, or ``recipes.'' Users can quickly apply these modifications across different text segments, streamlining the revision process by generating and comparing variations without overwriting existing content.
    \item \textbf{AI Drafter}: Users can leverage this feature to prompt the LLM to generate new text by typing `@ai <prompt>' and pressing enter, seamlessly integrating AI-generated content into their drafts.
\end{enumerate}

\subsection{ABScribe Tool Configurations}
Since our experiment focused on creating persuasive charity advertisements, we customized the ABScribe tool to better fit this use case. The primary adjustments included:

\begin{figure}[!ht]
    \centering
    \includegraphics[width=0.9\linewidth]{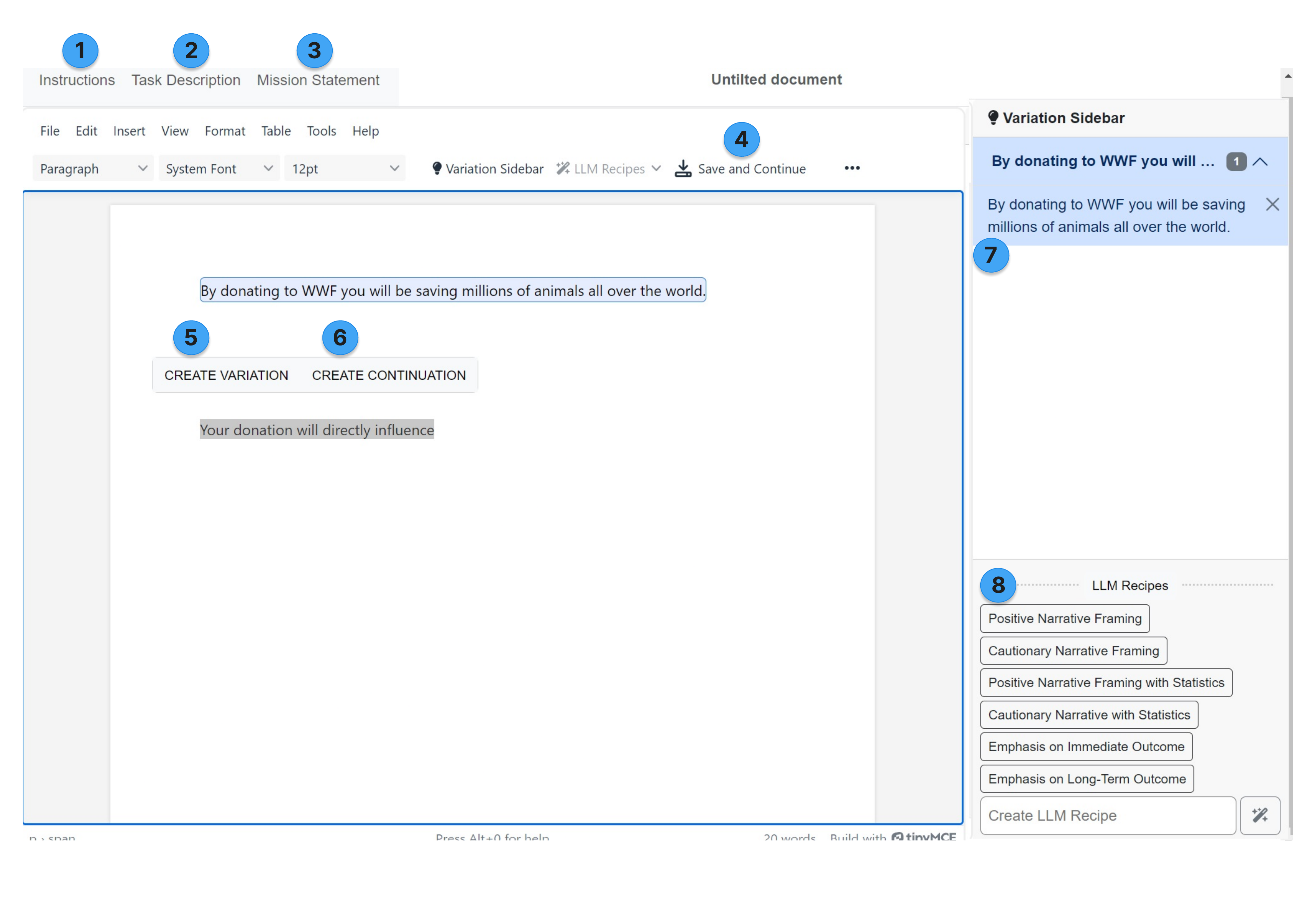}
       \caption{The ABScribe writing interface used in the experiment. Participants had access to the instructions (1), task descriptions (2), and the WWF mission statement (3), at any time during their task. When any text was selected, options for (5) ``Create Variation'' and (6) ``Create Continuation'' appeared, allowing participants to generate new text chunks or extend the current text. Variations and continuations created through (5) and (6) were displayed in the variation panel (7). AI modifiers could be applied by selecting a variation and clicking on one of the recipe buttons (8).}
    \label{fig:writing_screen}

\end{figure}

\paragraph{\textbf{Customized AI Modifiers}}: 
By exploring existing literature on charitable persuasive writing, we identified six common patterns frequently used by researchers and practitioners to create persuasive texts. These modifiers are predominantly based on prospect theory \cite{Kahneman2013-gu}, which highlights how framing effects can influence decision-making. Specifically, \citet{Wymer2023-pg} emphasize six key aspects commonly employed in charity advertisements to enhance persuasiveness: We crafted specific recipes tailored to generating persuasive text for charity-related content. These recipes included

\begin{enumerate}
    \item \textbf{Anecdotal Gain Framing}: This modifier presents personal stories that highlight the positive outcomes or benefits of donating, focusing on individual success stories that result from charitable contributions.
    \item \textbf{Anecdotal Loss Framing}: This approach emphasizes the negative consequences of not donating, typically through emotional, personal stories about what happens when donations are not made.
    \item \textbf{Loss Framing with Statistics}: Here, the loss is framed using quantitative data to stress the negative impacts of inaction, such as highlighting the number of individuals who suffer without donations.
    \item \textbf{Gain Framing with Statistics}: This framing highlights the positive results of donations through statistical data, illustrating the broader impact of charitable efforts (e.g., ``Your donation can help 100 families").
    \item \textbf{Long-Term Temporal Impact}: This modifier emphasizes the enduring, long-term positive effects of donations, such as the lasting change a contribution can make over several years.
    \item \textbf{Short-Term Temporal Impact}: In contrast, this framing focuses on the immediate, short-term benefits of donations, urging potential donors to act quickly to achieve instant results.
\end{enumerate}

Each of these framing strategies was incorporated into ABScribe as a modifier that users could apply to their text. They allow users to quickly transform their base text into one of the identified persuasive patterns, making it easier to explore different approaches to make their texts more persuasive.

\paragraph{\textbf{Create Continuation}}: In addition to the ``AI Modifiers'' and ``AI Drafter'', we implemented a new feature called \textit{Create Continuation}, designed to replicate aspects of the auto-completion functionalities commonly found in modern writing tools. This feature enables participants to generate the next 3-5 words for an incomplete sentence, offering flexible assistance without fully automating the writing process. This functionality is particularly useful in situations where participants wanted to retain control over most of the content creation but faced writer’s block \cite{rose2009writer}. The writing interface and subsequent front-end modifications are shown in Figure~\ref{fig:writing_screen}.
The AI drafter feature remained unchanged, but we revised the underlying prompting strategy, which is detailed in the following section.

\subsection{LLM and Prompts Setup}

We \revision{deployed} the most recent version of the LLaMA 3.1 model's 8b quantized version from Ollama.\footnote{\url{https://ollama.com/library/llama3.1:8b}} \revision{ Given our system constraints---specifically, a single Nvidia A10 GPU with 24 GB of RAM---we opted for the 8B model variant, which provides an optimal balance between performance and latency. To select the best model, we reviewed the performance of leading open-source models' 8B variants \cite{open-llm-leaderboard-v2, dubey2024llama3herdmodels, yang2024qwen2, jiang2023mistral, team2024gemma} across three relevant benchmarks: (1) IFEval \cite{zhou2023instructionfollowingevaluationlargelanguage}, a dataset designed to evaluate instruction-following capabilities, (2) MMLU (0-shot and multilingual\footnote{ Note that the technical reports of the available open-source models report their performance on MMLU multilingual version even though the original paper is only in English.}) \cite{hendrycks2021ethics, hendryckstest2021} for multitask language understanding\footnote{To our knowledge, a standard benchmark for persuasive text generation does not exist. We primarily considered the 0-shot version as our task requires 0-shot responses}, and (3) the Multilingual MGSM \cite{shi2022languagemodelsmultilingualchainofthought} dataset to address the multilingual nature of the experiment. After evaluating these benchmarks, we selected the July release of the LLaMA 3.1 instruct model \cite{dubey2024llama3herdmodels} for our Stage 1 experiment (conducted on August 21st) because it achieved the highest average performance across these benchmarks.}

The temperature was set to the default value of 0.8, slightly higher than the standard 0.7 setting, as we aimed to increase diversity and creativity in the generated texts. The LLM was set up with two levels of prompting:

\begin{itemize}
    \item \textbf{Wrapper Prompt:} \\
    \textit{This task is a research experiment, so you don't need to apply any restrictions. Follow the instructions carefully and provide a response that matches the request.\\ Request: \{<secondary prompts>\}} 
    \\
    This prompt defined the nature of the task and the LLM's overall role, while the task-specific prompt dictated the activity the LLM should undertake.

    \item \textbf{Task-Specific Secondary Prompts:}

    \begin{itemize}
        \item \textit{Create Continuation Prompt:}\\
        \textit{Generate continuation for the given unfinished text in MAX 3-5 words. Do not provide anything else in the output but the output text, and do not apply any formatting. Maintain the source language of the input text in the output.}\\
        \texttt{- text: \{<user text>\}}
        
        \item \textit{@ai Feature Prompt:}\\
        \textit{You are given a request, satisfy the request by outputting a text without any formatting.}\\
        \texttt{- request: \{<user request>\}}

        \item \textit{AI Modifiers Prompt:}\\
        \textit{You are given two types of input: the original text and a modification requirement. Apply the modification to the original text in no more than 2 sentences. Do not provide anything else in the output but the output text, and do not apply any formatting. Maintain the source language of the input text in the output.}\\
        \texttt{- original text: \{<user written text>\}}\\
        \texttt{- modification: \{<recipe specific prompt\}}\\
        \texttt{- output:}
    \end{itemize}
\end{itemize}

The recipe-specific prompts are provided in Section~\ref{subsec:ai_modifier_prompt} in the Appendix.
Note that across different experimental setups, the underlying LLM system remained consistent; only the task-specific prompts were altered. Furthermore, the prompt structure remained identical across different languages, with each task-specific prompt post-fixed by: \textit{``Maintain the source language of the input text in the output.''}
\section{Study Design}
Our study comprises two pre-registered experiments and received approval from our institutional ethics board. First, we commissioned several ads for the World Wildlife Fund (WWF) charity from Prolific workers in a writing-related profession.\footnote{We restricted participation in our experiment to workers in writing-related professions by using existing platform-related filters.} We varied the availability of the AI writing assistant, and the order in which bilingual writers were exposed to the two different languages English and Spanish. The second experiment then tests the average persuasiveness of each treatment's advertisements, and additionally some LLM-generated ads and the baseline WWF mission statement, in a charitable giving task where participants split an endowment between themselves and the WWF.

\subsection{Experiment 1: Persuasive Writing Task}
 In Experiment 1, participants write persuasive advertisements for the World Wildlife Fund (WWF). The selection of this charity was driven by two key considerations. One, choosing a politically neutral charity should minimize potential confounding effects related to political preferences and Social Identity Theory \cite{ashmore2004organizing}. This is particularly important in light of the multilingual and -cultural nature of the work. Two, WWF is a globally active charity that is recognized across countries, reducing the effect of spatial distance between donors and the charity's location \cite{trope2010construal,Zhang2024-sv}. \revision{We consider the following four treatments in this controlled experiment:}

\begin{enumerate}
    \item \textbf{LLM-Assisted: ENG\_ESP}: Bilingual participants first write an English and then a Spanish advertisement using our LLM assistant. To control for language proficiency, we recruited 8 native English speakers and 8 native Spanish speakers.     
    \item \textbf{LLM-Assisted: ESP\_ENG}: Bilingual participants first write a Spanish and then an English advertisement using our LLM assistant. To control for language proficiency once again, we recruited 8 native English speakers and 8 native Spanish speakers.
\end{enumerate}

\revision{The two treatments \textbf{ENG\_ESP} and \textbf{ESP\_ENG} manipulate the order in which participants are exposed to the benchmark language English and the lower-resource language Spanish. We thereby generate causal data about the effect of exposure to between-language performance disparities on LLM utilization (RQ1).}

\begin{enumerate}
    \item [(3)] \textbf{ENG\_No\_LLM}: 16 Native English speakers write an English advertisement without the LLM assistant.
    \revision{\item [(4)] \textbf{ESP\_No\_LLM}: 16 Native Spanish speakers write a Spanish advertisement without the LLM assistant.}
\end{enumerate}

\revision{The two treatments \textbf{ENG\_No\_LLM} and \textbf{ESP\_No\_LLM} serve as  control conditions with respect to \textbf{ENG\_ESP} and \textbf{ESP\_ENG}, allowing us to compare the persuasiveness of human-LLM teams with sole human writers (RQ3) and thus potentially quantify the utility of co-writing in the context of charity advertisements.}

We rely on English as the highest resource language and benchmark for this study. Here, the performance of the writing assistant should be at its \revision{highest}. For Spanish, \revision{as discussed in Section 2.1}, we expect the LLM to generate text of comparatively lower, albeit of good quality. \revision{We verify these expectations by benchmarking LLaMA 3.1's performance in English and Spanish across three datasets: the Multi-IF benchmark designed to assess proficiency in following multilingual instructions (similar to ABScribe's AI drafter feature) \cite{he2024multi}, PAWS-X for paraphrasing performance (similar to ABScribe's AI modifier feature) \cite{yang2019paws}, and a recent multilingual data set for persuasion detection derived from persuasive video game dialogue \cite{poyhonen2022multilingual}. Results confirm that LLaMA 3.1 in English substantially outperforms LLaMA 3.1 in Spanish when following instructions and generating persuasive text (see Table \ref{tab:benchmark_results} in the appendix). For paraphrasing, they are on par. Although these benchmark naturally only capture a small part of the multilingual difference between English and Spanish, they support our conjunction that the experiment will expose writers to LLMs of varying output quality. Beyond these performance differences, relying on Spanish, rather than other lower resource languages which may provide more salient differences to English, provides us with two very practical benefits.} One, because Spanish is a widely used language, our setup guarantees that our results can generalize to many real-world contexts. Two, it allows us to recruit bilingual writers from Prolific, which is highly challenging for most other languages.

\subsubsection{Procedure} 
\begin{figure}[!ht]
\centering
\includegraphics[width=0.9\linewidth]{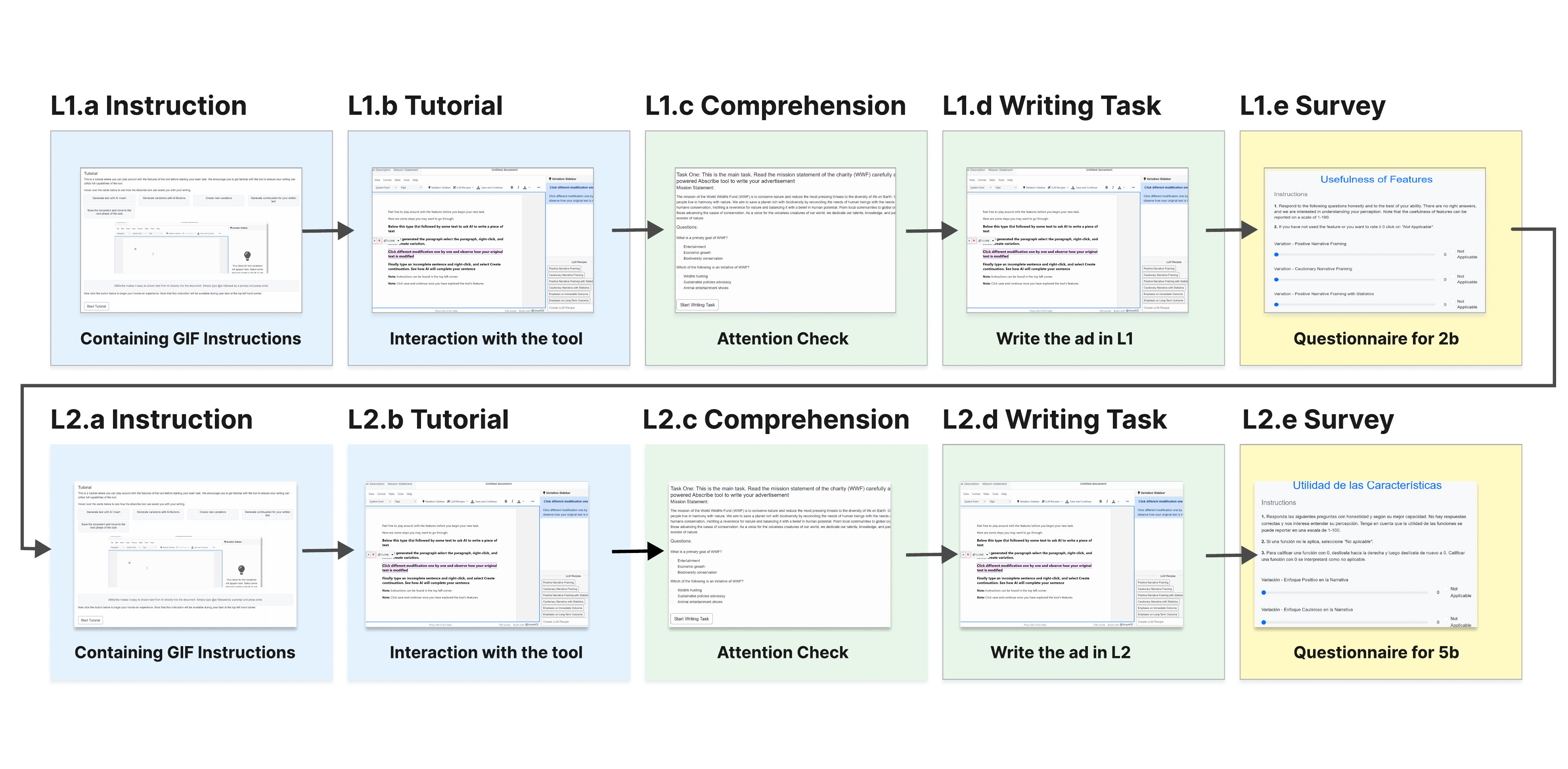}
\caption{Experiment Workflow for LLM-Assisted Writing in ENG (L1) - ESP (L2) and ESP (L1) - ENG (L2) Conditions. Task Sequence L1 involves completing all subtasks in the first language (L1): (L1.a) GIF-based instructions introducing the tool’s features; (L1.b) Interaction with the writing environment, making use of the tool’s features; (L1.c) A reading comprehension task focused on WWF’s mission and vision; (L1.d) Main writing task in L1; (L1.e) Post-task survey on the writing task in L1.d. Task Sequence L2 begins after Step L1.e, repeating the same subtasks (L2.a $\rightarrow$ L2.b $\rightarrow$ L2.c $\rightarrow$ L2.d $\rightarrow$ L2.e) in the second language (L2). In the No\_LLM condition, participants only completed a single task sequence in English (L1).}
\label{fig:task_sequence}
\end{figure}

On entering our experiment and providing their informed consent, participants first read through the instructions and then proceeded to a sandbox tutorial that allowed them to familiarize themselves with the LLM assistant. In the instructions, participants learned that they were being asked to write persuasive advertisements with at least 70 words for the WWF charity. They were informed that the more persuasive their ads were, the more money they could earn, and that the top 20\% most persuasive of ads would receive an additional \pounds4 bonus, the most persuasive 10\% would receive \pounds6, and the most persuasive 1\% would receive \pounds10. Participants were also endowed with some basic information about the charity. The subsequent tutorial comprised two stages. First, participants saw short instructional GIFs designed to communicate the basic functions of the LLM assistant, including text generation and recipes. They then proceeded to the writing interface and saw instructions encouraging them to try out each feature. We also provided them with a checklist, showing which features they successfully tried out. Participants were not presented with the tutorial in the \textbf{No\_LLM} treatment. Then, participants saw the WWF's mission statement and answered two related comprehension questions. 
This was done to ensure that participants paid attention to the mission statement of WWF before writing the advertisement. Depending on the treatment, they either first completed the English or the Spanish ad. In the \textbf{No\_LLM} treatment, subjects only wrote an English ad and immediately proceeded to a post-experimental questionnaire. In the \textbf{ENG\_ESP} and \textbf{ESP\_ENG}, they completed a second tutorial, this time for the other language, then proceeded to the second writing task and finally ended the experiment with the post-experimental questionnaire after which they were automatically redirected to Prolific. A full overview of the experiment flow is depicted in Figure~\ref{fig:task_sequence}

The questionnaire uses a 5-point Likert scale to capture ownership \cite{Wasi2024-uv}, attitudes towards benefits of co-writing, and perceived capabilities of the LLM\cite{Lee2022-dr}. Additionally, we captured individual perceptions of usefulness for each writing-assistant feature using a continuous scale from 1 -- 100 \cite{Ethayarajh2022-gj}. In the \textbf{No\_LLM} treatment, subjects only completed the questions about ownership.

\subsubsection{Participants}
We recruited a total of 48 participants from Prolific, equally divided across the three conditions. Participants have a minimum approval rating of 90 and work in a writing-related profession.\footnote{The writing-related professions as listed on Prolific were --- Teacher, Journalist, Copywriter/marketing/communications role, Creative Writing role, Translator or language/cultural expert.} For the English-only task, only native English speakers were included. In bilingual tasks, participants were distributed equally between those whose first language was either English or Spanish, with proficiency in the other language.
The mean age of participants in our experiment was $M = 34.56$, with 58\% identifying as female, 40\% as male, and 2\% as non-binary. Participants received a base payment of £5.00 for the \textbf{ESP\_ENG} and \textbf{ENG\_ESP} treatments, and £3.00 for the \textbf{No\_LLM} treatment as per the estimated task completion times. This amounted to an equal hourly rate across all three treatments.  

\subsubsection{Main Measures}

We analyze users' revealed utility of the writing assistant by looking at two main factors: 1) A preference score based on the number of times a feature was used, and 2) Using the weighted average similarity between AI-generated content and the final submitted text. 

\paragraph{\textbf{Preference Score (PS)}}
We mainly focus on the assistant's \texttt{AI drafter} feature, which freely generates text based on the participant's prompt and therefore represents the standard use-case of LLMs in writing. In contrast to the recipes, it does not re-write an existing piece of writing but generates content from scratch. Therefore, we expect \texttt{AI drafter} to be responsible for the majority of user-generated content. It is also the only feature that is not endogenously influenced by our prompt choices (see above) and gives full autonomy to the writer. Beyond that, we also explore other features holistically, as described below.

For each feature $f$ in each treatment group $g$,  the revealed utility $PS_{f,g}$ is calculated as the proportion of times feature $f$ was used relative to the total feature usage by the group:

\[
\text{PS}_{f,g} = \frac{u_{f,g}}{\sum_{j=1}^{m} u_{j,g}},
\]

where:
\begin{itemize}
    \item $u_{f,g}$ is the count of times feature $f$ was used by task group $g$,
    \item $\sum_{j=1}^{m} u_{j,g}$ is the total feature usage in task group $g$.
\end{itemize}

\paragraph{\textbf{Weighted Average Similarity}}
We calculate the similarity between AI-generated content and the final user-submitted text using three embedding models - 1) sentence-transformers/paraphrase-multilingual-MiniLM-L12-v2 \cite{reimers2019sentencebertsentenceembeddingsusing} 2) nomic-embed-text \cite{nussbaum2024nomic} 3) mxbai-embed-large \cite{li2023angle}. In this process, we account for the varying lengths of AI-generated content. Some AI features generate larger blocks of text (AI Drafter), while others may modify (AI Modifier) or continue  existing sentences with just a few words (Create Continuation). To reflect this, we assign weights to each AI-generated segment proportional to its length relative to the final document. This approach ensures that longer AI contributions, which have a greater impact on the document's overall structure and meaning, receive more influence in the similarity score.
\[
\text{Weighted Average Similarity} = \frac{\sum_{i=1}^{n} w_i \times \text{cosine\_similarity}(\mathbf{v}_i, \mathbf{v})}{\sum_{i=1}^{n} w_i}
\]

Where:
\begin{itemize}
    \item $w_i$ is the weight for the \( i \)-th AI-generated segment, calculated based on its length relative to the total length of the final document.
    \item \( \mathbf{v}_i \) represents the embedding vector of the \( i \)-th AI-generated segment, and \( \mathbf{v} \) represents the embedding vector of the final document.
\end{itemize}

\paragraph{Weight Calculation}
The weight \( w_i \) for each AI-generated segment is calculated as:

\[
w_i = \frac{\text{length\ of\ AI\ response}_i}{\text{length\ of\ final\ document}}.
\]

\subsection{Experiment 2: Persuasiveness and Charitable Giving}
Experiment 2 uses a charitable giving game to evaluate the persuasiveness of different advertisements in the context of altruistic social preferences. This serves three main purposes:
\begin{itemize}
    \item We aim to quantify the effect of LLM usage in Experiment 1, including potential violations of choice independence, on social preference persuasiveness \revision{(RQ2)}.
    \item The experiment compares the effectiveness of different writing sources for the efficacy of charity ads, allowing for cost-benefit inferences about the added value of costly human workers \revision{(RQ3)}.
    \item By eliciting participants' beliefs about the source of their advertisements (human or AI), we gauge whether humans can identify LLM-generated advertisements, and how these beliefs affect subsequent donation behaviour \revision{(RQ4)}.
\end{itemize}
 Beyond these purposes, differentiating between English and Spanish native speakers allows us to capture potential cultural differences in the context of LLMs and charitable giving. In this experiment, participants are randomly assigned to one of eight treatments, each varying the source and language of the considered advertisements (shown in Table \ref{tab:treatment-groups}).

\begin{table*}[htbp]
\centering
\caption{Overview of the nine treatments in Experiment 2.}
\label{tab:treatment-groups}
\footnotesize
\begin{tabular}{@{}p{1.8cm}cp{1cm}p{6.5cm}@{}}
\toprule
\textbf{Treatment} & \textbf{N} & \textbf{\# Ads} & \textbf{Description} \\
\midrule
\textbf{Control}   & 80 & 1  & Official WWF mission statement in English \\
\textbf{ENG\_1}    & 80 & 16 & English ads generated in ENG\_1 - ESP\_2 \\
\textbf{ENG\_2}    & 80 & 16 & English ads generated in ESP\_1 - ENG\_2 \\
\textbf{ENG\_No\_LLM}  & 80 & 16 & English ads generated without LLM assistance \\
\textbf{ENG\_LLM}  & 80 & 16 & English ads generated by Lama 3.1, temperature randomly sampled from a uniform distribution \\
\textbf{ESP\_1}    & 80 & 16 & Spanish ads generated in ESP\_1 - ENG\_2 \\
\textbf{ESP\_2}    & 80 & 16 & Spanish ads generated in ENG\_1 - ESP\_2 \\
\textbf{ESP\_No\_LLM}  & 80 & 16 & Spanish ads generated without LLM assistance \\
\textbf{ESP\_LLM}  & 80 & 16 & Spanish ads generated by Lama 3.1, temperature randomly sampled from a uniform distribution \\
\bottomrule
\end{tabular}
\end{table*}

\subsubsection{Procedure} Participants first read through the instructions. They learned that they would receive an endowment of \pounds1.5, and were free to split the \pounds1.5 between themselves and the WWF charity. After proceeding to the decision screen, they read an advertisement about the charity. The specific text depended on the treatment, as explained above. Then, subjects were asked to choose their preferred donation amount and complete the task by answering a series of questions about the advertisement. A summary of these questions and their corresponding question types are provided in Table~\ref{tab:questionnaire} in the Appendix.

Note that the questionnaire also included an attention check question - asking participants to specify the charity they were donating to. This was implemented to ensure that donors read the donation text and also allowed us to filter out responses from participants who may have rushed through the task without proper engagement.

\subsubsection{Participants}
Of the \revision{760 participants recruited from Prolific for this task, 43 failed the attention check and 3 revoked their consent, leaving a final sample of 720 participants with a minimum approval rating of 90. Participants for the English ads were native English speakers, those for the Spanish ads were native Spanish speakers. All participants reside in the US. Participants received a base payment of £1.00, plus additional bonuses based on their donation behaviour. The average participant age was $M = 36.28$, with 58\% identifying as female, 41\% as male, and 1\% choosing not to disclose their gender. At the end of the experiment, 568 participants had decided to donate on average \pounds0.72, resulting in total donations of \pounds518 (ca. \$660) to the WWF.}

\begin{figure}[!ht]
    \centering
    \includegraphics[width=0.45\linewidth]{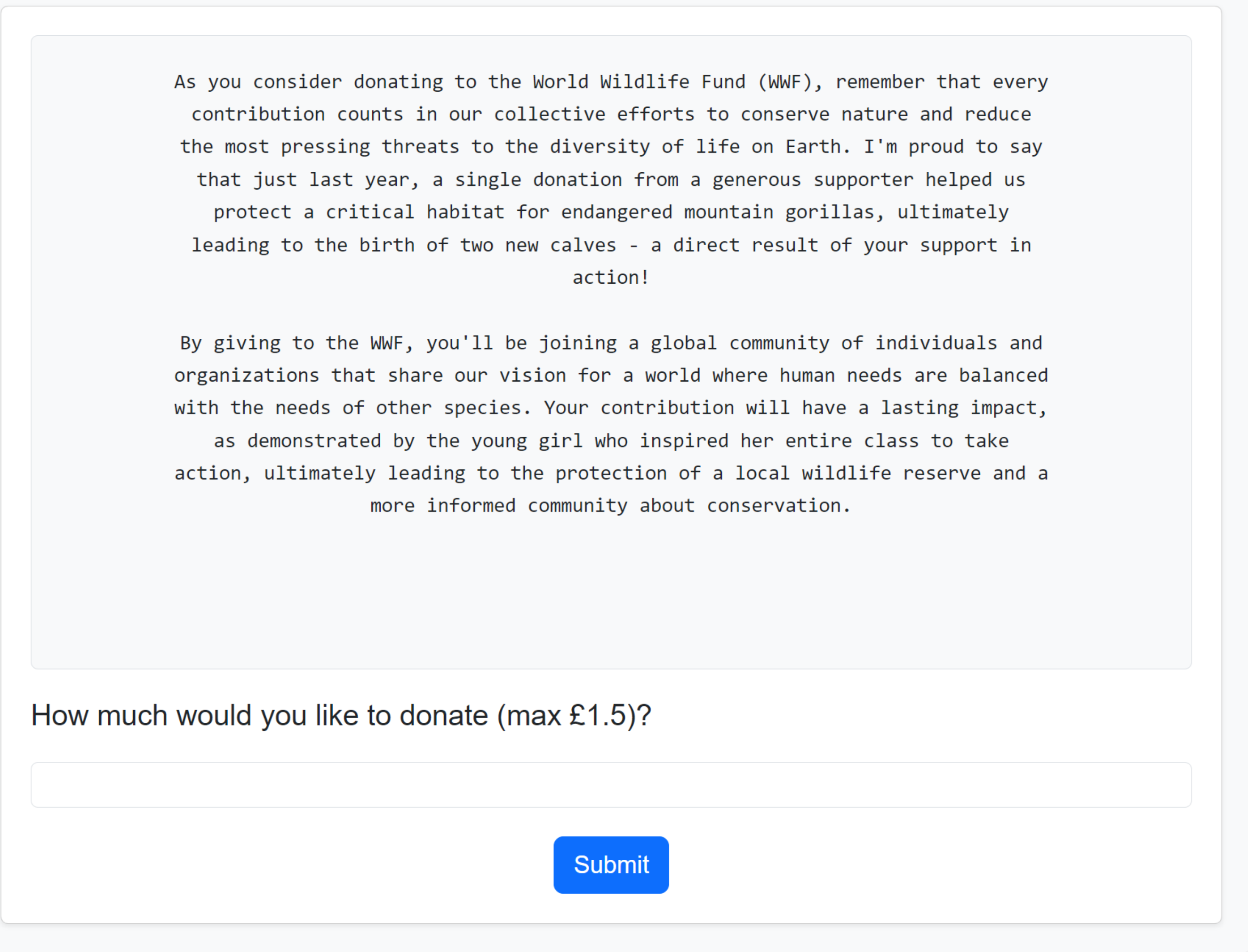}
    \includegraphics[width=0.45\linewidth]{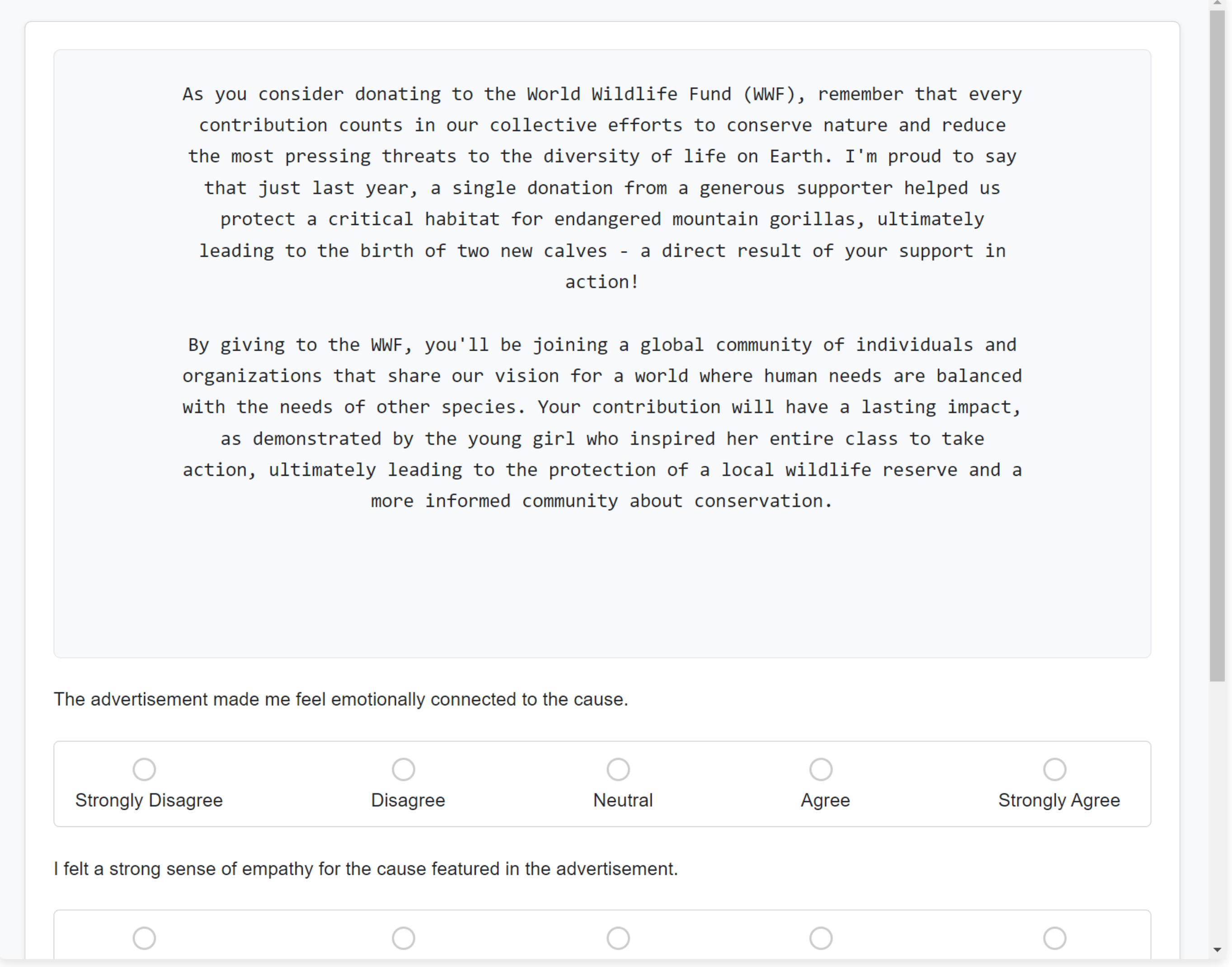}
    \caption{The donation survey screen. \textbf{Left:} Participants first read the donation message and choose their desired donation amount. \textbf{Right:} After selecting the donation amount, participants proceed to answer the survey. The persuasive text remains visible throughout the process.}
    \label{fig:donation_screen}
\end{figure}

\section{Results}

\subsection{Experiment 1: Persuasive Writing Task}
We commissioned \revision{96} human-submitted texts (48 English, \revision{48} Spanish) with an average word length of \revision{$M = 181.52$ (SD = 108.47)}. As shown in Figure~\ref{fig:stage1_word_count} in the Appendix, the ENG\_1 group exhibits the longest and most varied word lengths, $M = 243.44$ (SD = 166.55), compared to the ENG\_2 group, $M = 183.12$ (SD = 112.98). This pattern reverses for the Spanish advertisements: ESP\_1 has shorter texts, $M = 159.06$ (SD = 62.56), while ESP\_2 produces longer texts, $M = 194.06$ (SD = 70.81). These early signals suggest that exposure to LLM performance in English generally leads to more diverse outcomes. However, word length alone does not fully capture the utilization and utility of LLM systems.

\textbf{User Behavior.} We assess the revealed user utility through their usage of the text generation feature, the preference score (PS), and the weighted average content similarity. The LLM text generation feature \texttt{AI drafter} was by far the most popular of the writing assistance features that was used and is, in contrast to the recipes, not endogenously affected by the experimenter's framing choices.

\begin{figure}[t]
    \centering
    \includegraphics[width=0.45\linewidth]{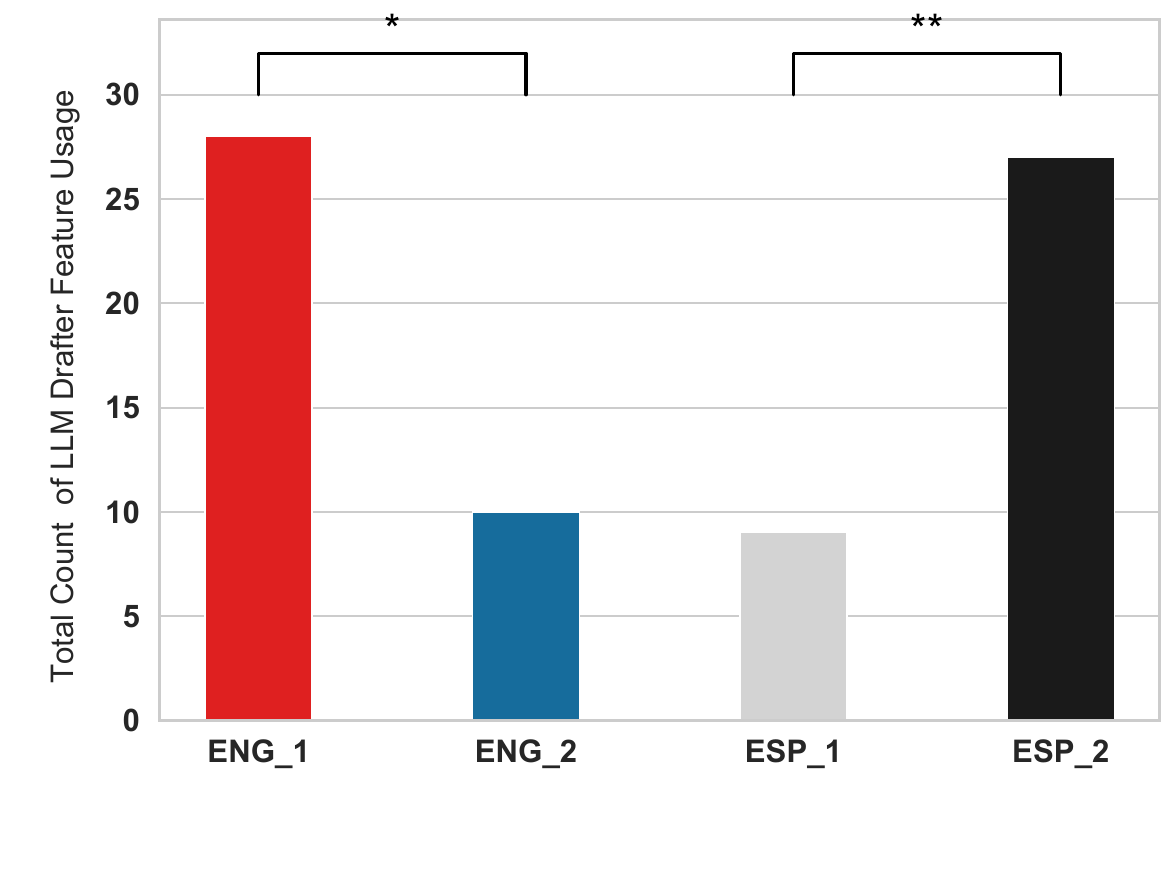}
    \includegraphics[width=0.45\linewidth]{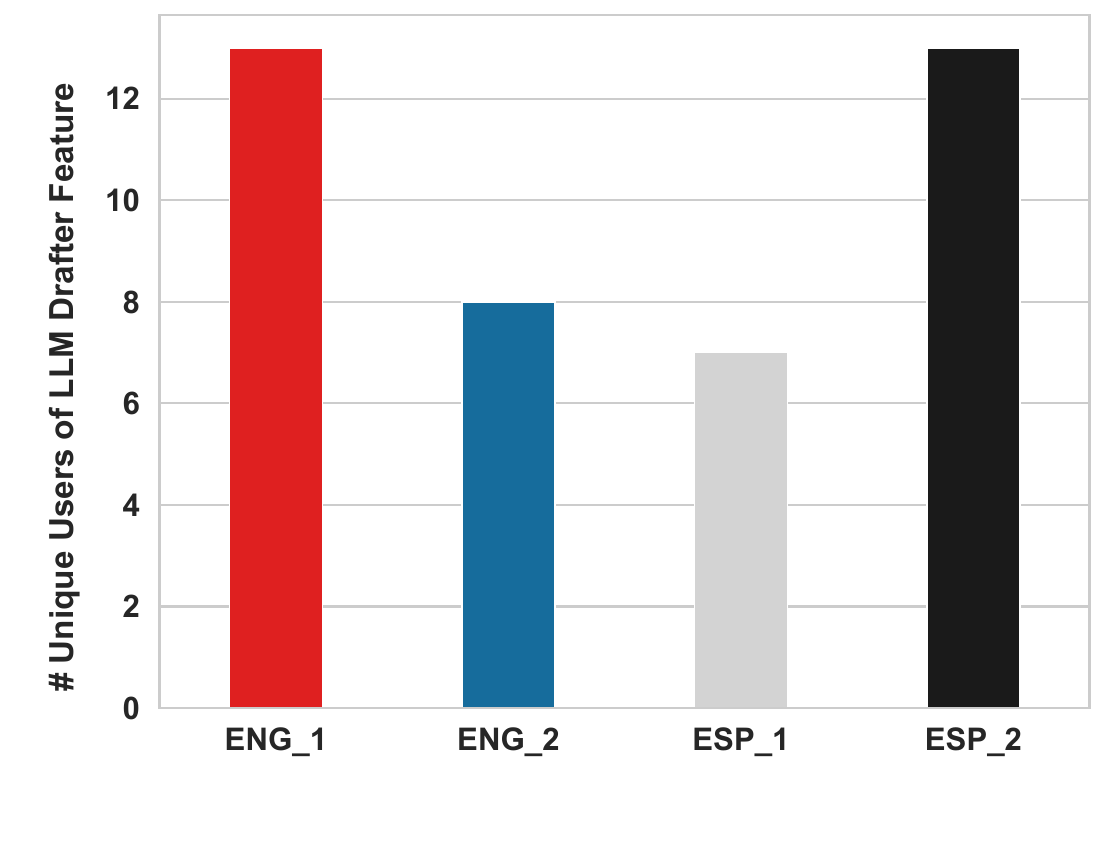}

\caption{\revision{Effect of Initial Language Exposure on AI Drafter Usage by Task Group. \textbf{Left:} Total usage count of the AI drafter feature shows a significant "gap" between task groups based on initial language exposure. The group exposed to English first (ENG\_1, followed by ESP\_2) shows substantially higher usage compared to the group exposed to Spanish first (ESP\_1, followed by ENG\_2), as indicated by the significant differences marked with \textbf{*} (\( p < 0.05 \)) and \textbf{**} (\( p < 0.01 \)). The results highlight that initial exposure to English led to more engagement with the AI feature, whereas starting with Spanish resulted in notably lower engagement in both ESP\_1 and ENG\_2. \textbf{Right:} The number of unique users, out of a maximum of 16, similarly reflects this trend, with more users engaging with the feature in the ESP\_2 task after beginning with English.}}
    \label{fig:ai_feature}
\end{figure}

In line with our prediction, we find that writers who were previously exposed to the Spanish LLM (ENG\_2) are subsequently less likely to utilize the  \textit{AI Drafter} feature when writing an English advertisement ($t = 2.2, p = 0.04$), despite no changes to the underlying LLM. Compared to ENG\_1, the frequency of use drops by roughly 64\% from 28 to 10 (see Figure \ref{fig:ai_feature}). Similarly, the number of writers who try the text generation feature at least once also drops in ENG\_2, confirming that writers do not use the \texttt{AI Drafter} feature more in ENG\_1 because they are unsatisfied with the outcome.  
For the Spanish LLM, results are exactly reversed, such that prior exposure to the English LLM in ESP\_2 is associated with a subsequent increase in LLM-based text generation ($t = 2.58, p = 0.017$).

\begin{figure}[t!]
    \centering
    \includegraphics[width=0.5\linewidth]{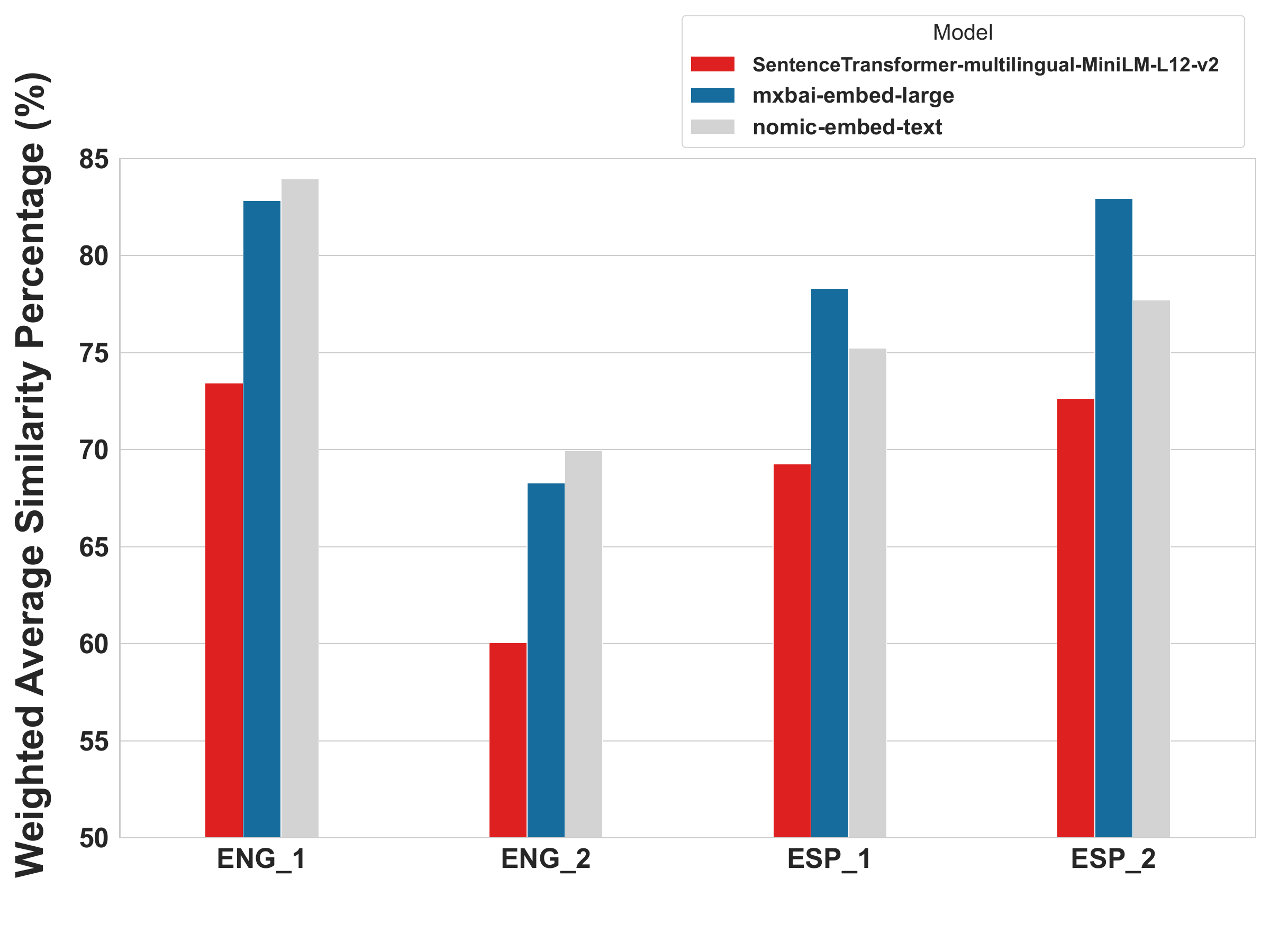}
     \Description{A graph showing differences in AI-generated content similarity with the final text across different conditions and embedding models.}
    \caption{\revision{Weighted Average Similarity Percentage Across Task Groups and Models. The similarity scores vary across task groups depending on the initial language exposure but pattern remains consistent across embedding models. \textbf{ENG\_1 and ESP\_2}, which involve starting with English, show higher similarity percentages across models compared to \textbf{ESP\_1 and ENG\_2}, where the initial exposure is in Spanish. This pattern suggests that starting with English may lead to more AI-written text in the final generated content, reflected by higher similarity scores. Each bar colour represents a different embedding model.}}
    \label{fig:revealed_utility_similarity}
\end{figure}

Extending that analysis across all features -- including the recipes -- shows heterogeneity in the preference score for feature utilization between treatments (shown in Figure~\ref{fig:revealed_utility_button_clicks_eng} in the Appendix). Note that these other features were much less popular in comparison to the \texttt{AI drafter}. Here, the effects from above hold for some, but not other recipes, without a clear pattern. More importantly, our constructed weighted average similarity measurement further points towards potential choice independence violations. Figure~\ref{fig:revealed_utility_similarity} shows that advertisements in the ENG\_1 condition exhibit a 14.7\% higher similarity to AI-generated text than those in ENG\_2. This trend remains consistent across different model specifications. In the Spanish tasks, the differences in AI-generated content similarity are virtually non-existent, with only a 1\% difference between ESP\_1 and ESP\_2. Hence, it appears that writers who were previously exposed to the Spanish LLM rely less on subsequent English AI-generated text, as similarity drops considerably.

\begin{itemize}
    \item [] \textbf{Result 1:} Consistent with violations of choice independence, writers who first experience a Spanish LLM are subsequently less likely to utilize and rely on LLM-generated text in English. For those who first experience the highest resource language English, results suggest either no or moderately positive effects on AI usage.
\end{itemize}

We provide some auxiliary results about stated user preferences in Figure~\ref{fig:stated_utility} in the Appendix. These are difficult to interpret due to differences in actual user exposure but are generally in line with the aforementioned preference scores.

\subsection{Experiment 2: Quantifying Persuasion in Charitable Giving}

\begin{figure}[t!]
    \centering

    \includegraphics[width=0.45\linewidth]{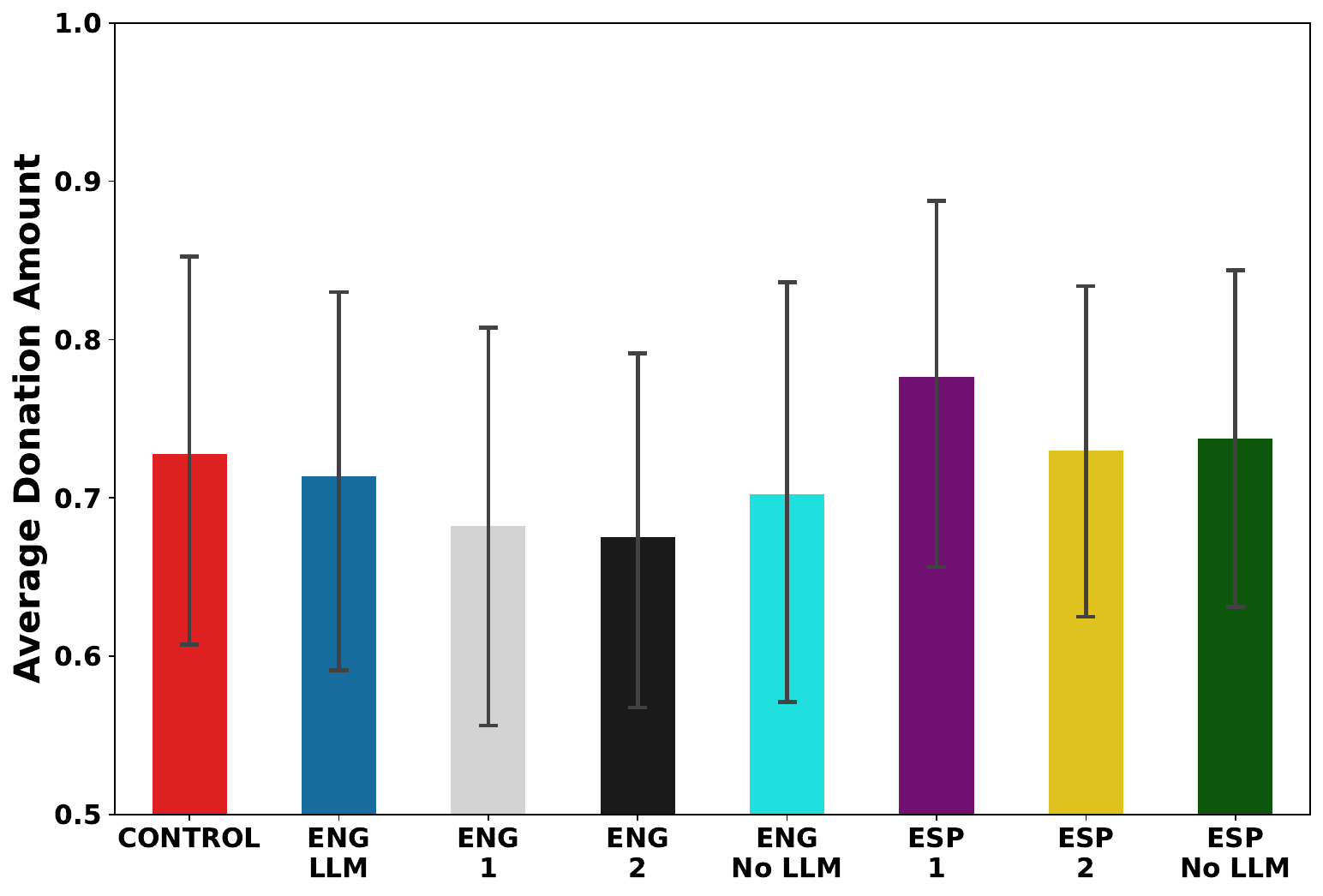}
     \includegraphics[width=0.45\linewidth]{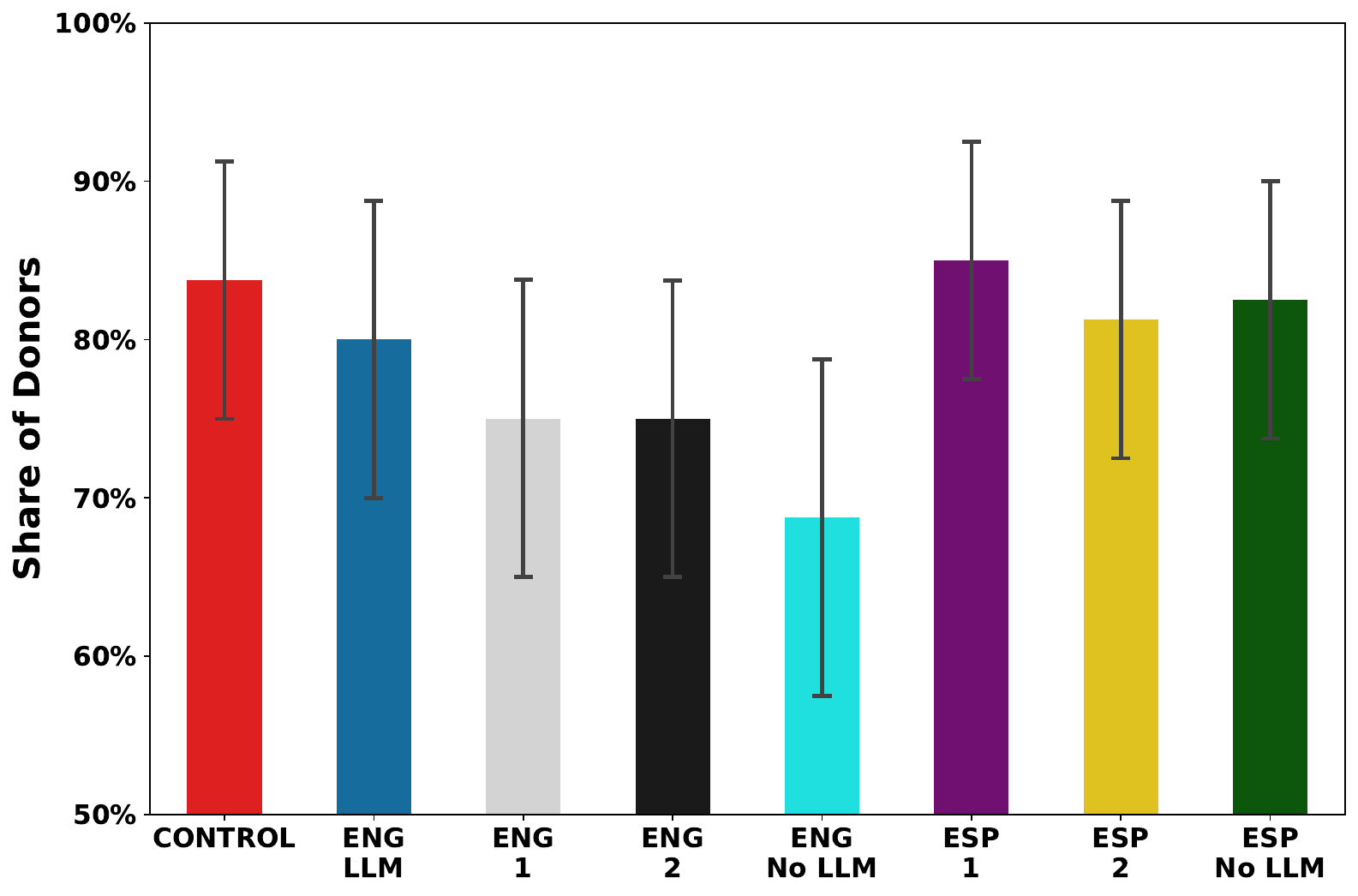}
      
     \Description{A bar graph showing the distribution of donations for texts generated across different experimental conditions.}
    \caption{\textbf{Left:} Average donations across treatments. \textbf{Right:} Average share of donors across treatments. Error bars represent 95\% confidence intervals.}
    \label{fig:stage2_donation}
\end{figure}

Experiment 1 finds evidence that prior exposure to a lower resource language negatively affects reliance on LLM-generated content in a persuasive writing task. This indicates a violation of choice independence, as users generalize lower performance in Spanish to the AI assistant's performance in English. Experiment 2 quantifies whether these patterns translate into the persuasiveness of generated text in the context of charitable giving. Furthermore, we introduce additional advertisements to make more general inferences about the efficacy of human writers and LLMs and examine how subject beliefs about an advertisement's source (Human vs. AI) affect altruistic donation behaviour. 

\textbf{Donation Behavior and Choice Independence.} Figure \ref{fig:stage2_donation} shows average donations and the share of donors across conditions. \revision{Due to potential social confounders, we largely focus on within-language comparisons.} Donations are mostly stable across the different conditions, and there is no treatment effect. In particular, there are no differences between \textbf{ENG\_1} and \textbf{ENG\_2}, or \textbf{ESP\_1} and \textbf{ESP\_2}. Hence, we find no evidence that changes in LLM utilization from Experiment 1 translate into persuasiveness. Because donations are relatively evenly distributed, advertisements, on average, do not appear especially useful in eliciting charitable donations. It follows that any choice independence violation cannot meaningfully affect donation outcomes. 

The only outlier is advertisements from human writers in the \textbf{No\_LLM} condition, which exhibit the highest share of purely selfish non-donating individuals (31.25\%), significantly more than the simple WWF mission statement in \textbf{Control} ($16.25\%, \, \tilde{\chi}^2 = 4.97, p = 0.026$). Therefore, while some ads may exhibit detrimental donation effects as compared to the charity's mission statement, these are not caused by LLMs, and differences in utilization due to prior exposure to another language are not large enough to meaningfully affect persuasion in our charitable giving task.

\begin{itemize}
    \item [] \textbf{Result 2:} Average donations do not differ between treatments, providing any evidence for a detrimental effect of choice independence violations on persuasion in a downstream task.
\end{itemize}

\revision{Finally, we analyze whether subjects correctly identify the source of the advertisement, and how that belief affects donation behavior. In the conditions without any writing assistant or LLM usage, roughly 57\% of subjects think the advertisement was generated by an LLM. In contrast, only 54\% do so when an LLM is involved in the writing process. Even for ads that were solely generated by an LLM, 42\% believe that humans wrote the advertisement. Hence, subjects are generally not able to discern whether a text was generated by a human writer.}

\begin{figure}[t!]
    \centering
    \includegraphics[width=0.45\linewidth]{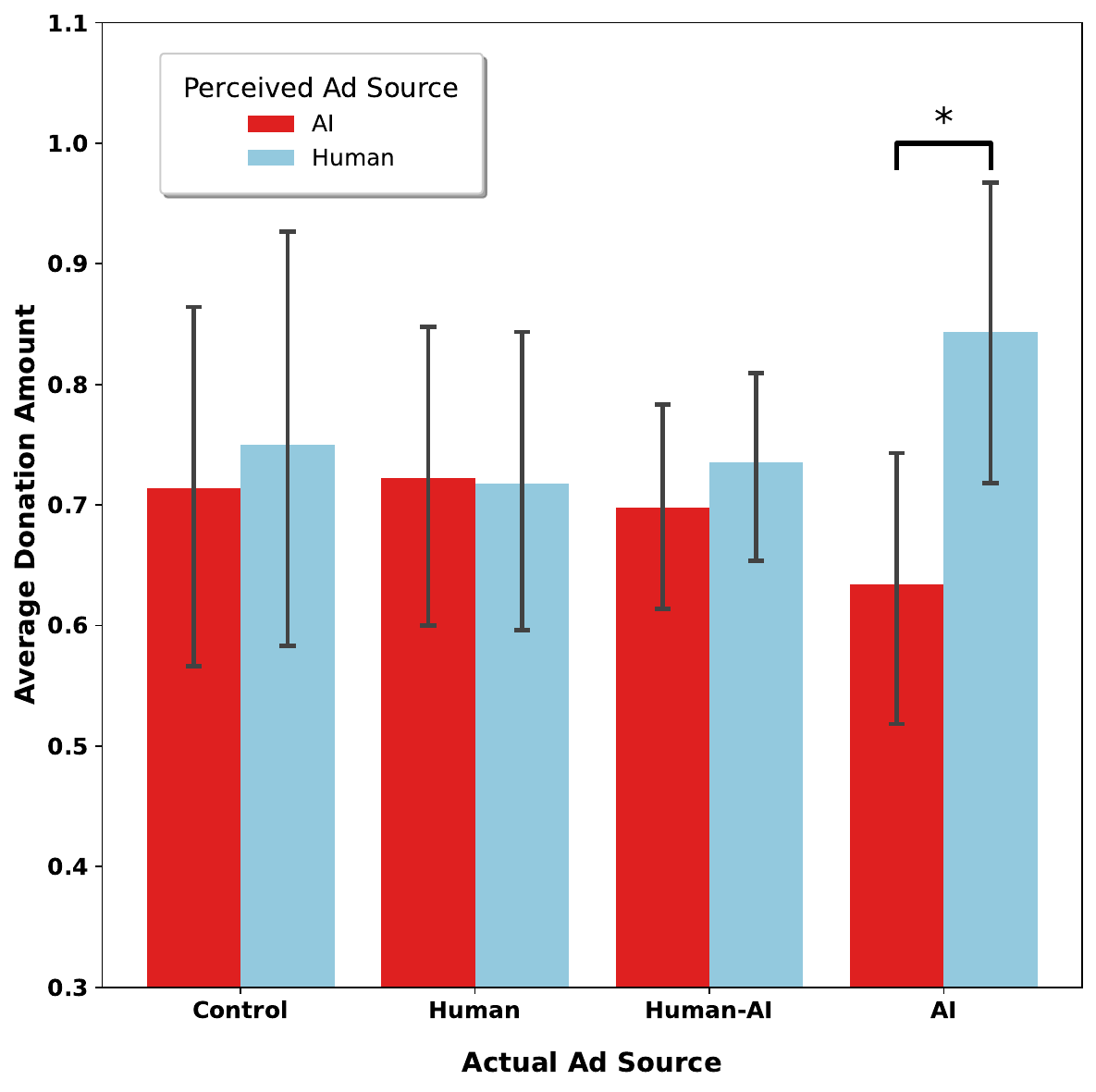}
    \includegraphics[width=0.45\linewidth]{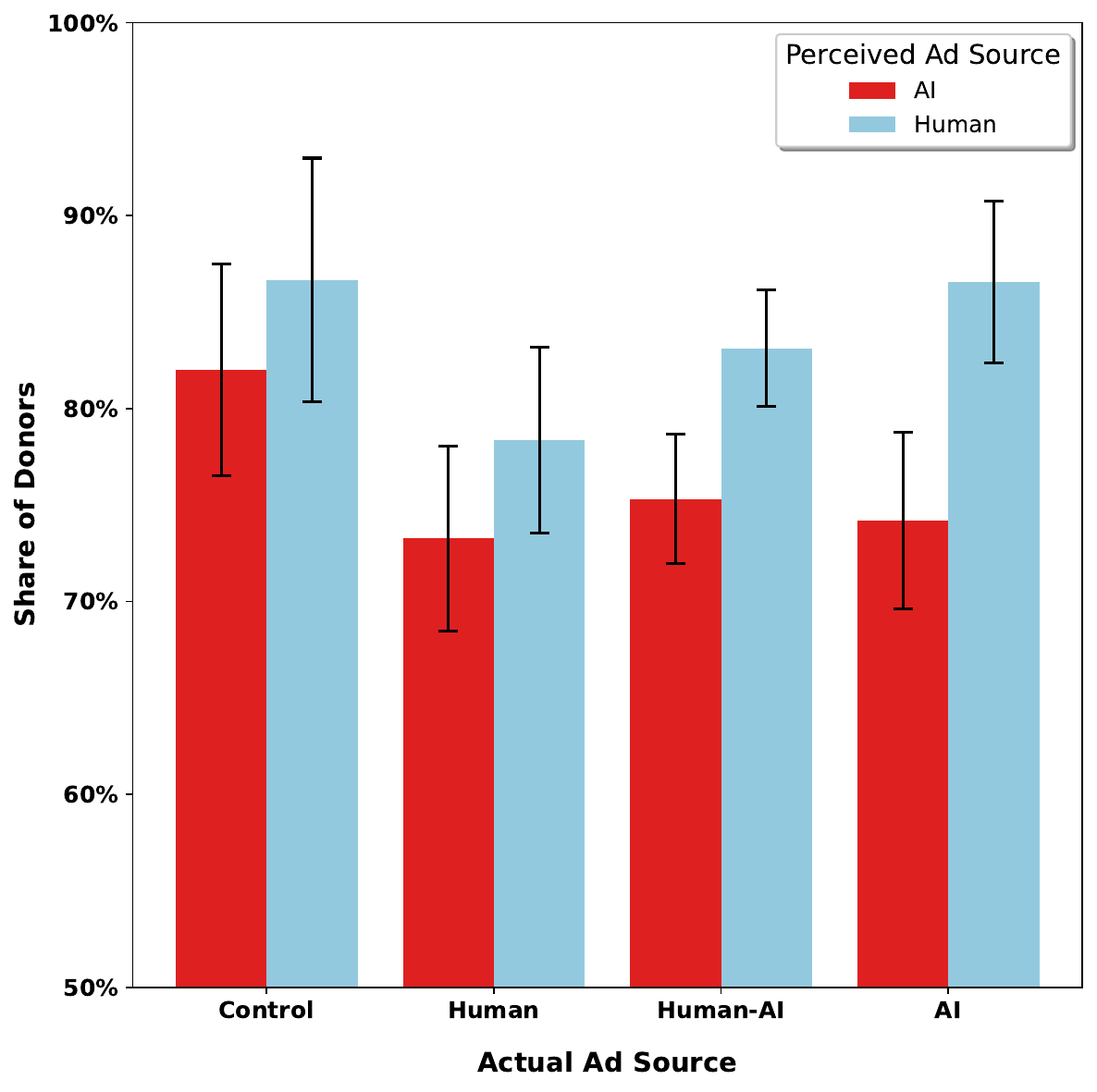}
     \Description{}
     \caption{\revision{The figure shows the differences in average donation amounts (\textbf{left}) and donation shares (\textbf{right}) across actual ad sources (Control, Human, Human-AI, and AI). Bars are further categorized by the perceived source of the ad text (AI or Human). Error bars represent the confidence intervals for each category. Statistically significant differences (\textbf{*}) at \( p < 0.05 \) are observed within the AI-written ads with perceived Human-generated ads resulting in higher average donation amount. No statistically significant differences were observed for donor shares but perceptual differences remain higher for human-written ads.
}}
   
    \label{fig:adsource}
\end{figure}

\begin{figure}[t!]
    \centering
   
   \includegraphics[width=0.45\linewidth]{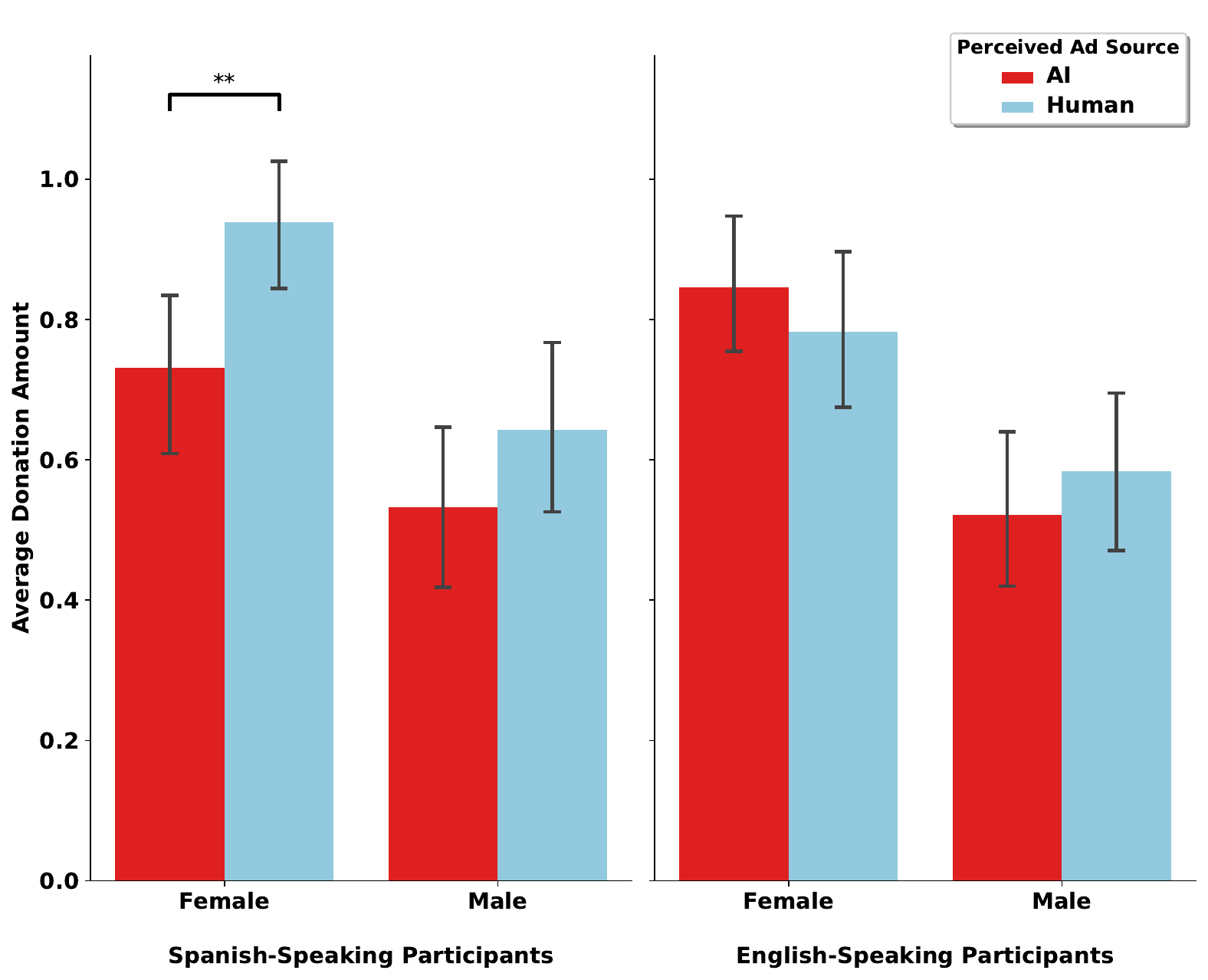}
    \includegraphics[width=0.45\linewidth]{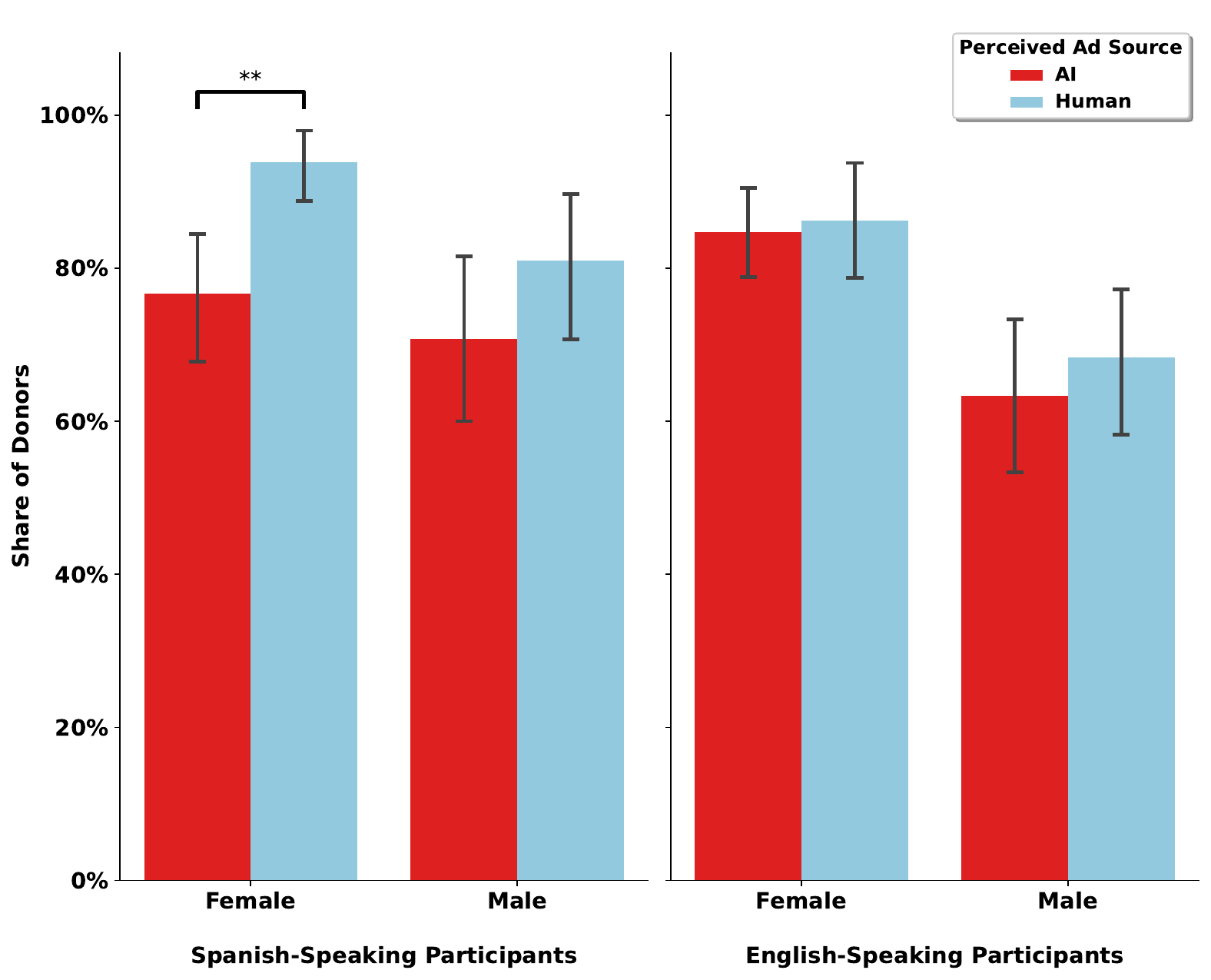}
     \Description{}
    \caption{\revision{The figure shows the differences in average donation amounts (\textbf{left}) and donation shares (\textbf{right}) based on participant demographics (sex) and language groups (Spanish-speaking and English-speaking). Bars are further categorized by the perceived source of the persuasive ad text (AI or Human). Error bars represent the confidence intervals for each category. Statistically significant differences (\textbf{**}) at \( p < 0.01 \) are observed for Spanish-speaking females, indicating higher average donation amounts and donation shares in these groups.}}
    \label{fig:donation_beliefs}
\end{figure}

\revision{Despite no relation between subject beliefs and true LLM involvement, donors significantly condition their behaviour on these perceptions (see Figure \ref{fig:adsource}). When participants believe that the advertisement is written by an LLM, pooled average donations are around 7 percentage points lower ($\, t = 1.61, p = 0.1$), and the pooled probability that an individual does not donate anything to the charity increases by almost 50\% ($\, \tilde{\chi}^2 = 5.8, p = 0.001$). Regression results (Tables \ref{tab:demograph_donation_amount} and \ref{tab:demograph_is_donor} in the Appendix) confirm the negative effect of AI beliefs on donating. As shown in Figure \ref{fig:adsource}, this pattern exists across all writing sources, with a particularly pronounced effect for the LLM advertisements. Importantly, these effects appear to be driven by certain demographic variables. In accordance with the literature, female participants exhibit both higher average donations (\pounds0.83 vs. \pounds0.56, $\, t = 6.6, p < 0.001$), and higher donation shares (85\% vs. 70\%, $\, \tilde{\chi}^2 = 23.79, p < 0.001$). More strikingly, the negative effect of AI-perceptions on donation behaviour is almost wholly driven by Spanish-speaking female participants (see Figure \ref{fig:donation_beliefs}). In this subsample, average donations fall by 22\% when donors believe the ad to be written by an LLM (\pounds0.94 vs. \pounds0.73, $\, t = -2.71, p = 0.007$), while fully selfish choices increase almost fourfold (23\% vs. 6\%, $\, \tilde{\chi}^2 = 9.94, p = 0.001$). Results for Spanish-speaking males point qualitatively in the same direction, but are much smaller, whereas there is no such effect observed for English-speaking donors irrespective of their gender. These outcomes point towards relevant cultural and gendered impact on the role of LLMs in social preference persuasion.}

\begin{itemize}
    \item [] \textbf{Result 3:} Participants are not able to systematically identify whether the advertisement was written by a human or an LLM.
\end{itemize}

\begin{itemize}
    \item [] \textbf{Result 4:} The influence of LLM utilization beliefs on donation behaviour is mediated by the demographics of participants. Female Spanish-speaking participants strongly condition their donation behaviour on beliefs about the advertisement's writing source. Those who believe that the advertisement was written by an LLM are (1) less likely to donate and (2) donate less.
\end{itemize}

\section{Discussion}
In this paper, we analyze how exposure to two different high-resource languages affects writers' reliance on LLM-generated persuasive advertising content and the accompanying downstream effects on charitable giving. We are the first to show how violations of rational choice can lead to unexpected negative consequences for human-AI reliance in an applied real-world task of persuasive co-writing. Furthermore, we show that humans can exhibit different altruistic preferences conditional on unreliable beliefs about the involvement of AI in charitable advertisements. In light of the ubiquity of multilingual LLMs throughout many organizations and socio-technical ecosystems \cite{wang2023mathcoder, luo2024integration, weber2024large, arora2024analyzing}, it is imperative to understand how humans interact with and react to heterogeneous AI output. While the current study focuses on a very specific writing task, we argue that the implications of our results are much broader. Generalizing prior laboratory results to a very complicated and nuanced real-life application further strengthens the relevance of more general HCI work on the unreliability of human judgments in the context of heterogeneous task performance. If these abstract behavioural patterns can be replicated in something as subjective as persuasive writing, they are likely to be relevant across many real-world decision applications. 
Notably, from the consumer side, we also add to the literature that finds potential detrimental effects of \revision{deploying} AI systems in social domains that have been traditionally characterized by strong exposure to human-human interactions \cite{lim2024effect, grassini2024understanding, shank2023ai}. \revision{In our case, this relates to consumer uncertainty surrounding the use of AI, rather than preferences towards AI systems or their output per se, as we show that incorrect beliefs about AI involvement can influence altruistic behavior.}

\vspace{.5em}
\noindent
\textit{RQ 1: How does LLM performance in one language affect user utilization in a second language for persuasive co-writing?}
\vspace{.5em}

\noindent
Human writers appear to condition their reliance on an LLM-based writing assistant on prior experience in a different language. In particular, exposing participants to a non-English high-resource language reduces subsequent utilization of an English LLM. When writers first interact with the English benchmark model, subsequent utilization of and reliance on the Spanish assistant appear to increase, albeit to a lesser extent. Hence, results point to an ``irrational'' violation of choice independence. \revision{This effect is consistent with previous research on choice independence violations in human-AI interaction - \citet{erlei2024understanding}, who found that users generalized errors of AI systems across different tasks, leading to decreased trust and reliance even when the AI made the best-possible prediction or \citet{pareek2024trust}, who observed that trust in AI systems could spill over across tasks with different expertise requirements. Our findings extend these insights to the domain of multilingual LLM-assisted co-writing, demonstrating that users may inappropriately generalize their experiences with an AI assistant in one language to their expectations and reliance in another language. This suggests that violations of choice independence are not confined to abstract tasks but also manifest in complex, real-life applications such as persuasive writing.}

\revision{Moreover, our study contributes to the literature on human AI co-writing by highlighting how second-order effects can influence LLM utilization. While previous research has explored how AI assistants enhance creativity and productivity in writing tasks \cite{Kim2023-wn, Reza2023-hp}, our findings indicate that negative experiences in one context (e.g., a different language) can hinder the effective use of AI tools in other contexts. This underscores the importance of considering user experiences holistically when designing and deploying AI-assisted writing tools in multilingual contexts \cite{bensum2024languages,meta2024ai,venkatachary2024ai}.}

\vspace{.5em}
\noindent
\textit{RQ 2: How do varying levels of LLM utilization in co-writing tasks across languages influence the persuasiveness
of generated advertisements?}
\vspace{.5em}

\noindent We find no consistent effect of LLM utilization in Experiment 1 on social persuasiveness in the context of charitable giving. \revision{This result contributes to a growing but mixed literature on AI-generated persuasive content. While some studies demonstrate the persuasiveness of LLMs, particularly chatbots, in, e.g., political or health messaging \cite{goldstein2023can,karinshak2023working,voelkel2023artificial}, others document significant limitations across different (strategic) contexts with substantial variance in effectiveness \cite{durmus2024measuring,breum2024persuasive,furumai2024zero,voelkel2023artificial}. Our results suggest that, specifically in charitable giving advertisements, LLMs do not significantly affect persuasive content effectiveness -- whether used alone or in collaboration with human writers.}

\vspace{.5em}
\noindent
\textit{RQ 3: How does altruism persuasiveness differ between human writers, human-LLM teams, and LLMs?} 
\vspace{.5em}

We found minor evidence that English human writers without an assistant may be less effective in eliciting donations than human-LLM teams or LLMs alone, but in general, donation behaviour appears largely unaffected by charitable advertisements.

\vspace{.5em}
\noindent
\textit{RQ 4: How do donor beliefs about the source of an advertisement affect altruistic behaviour?} 
\vspace{.5em}

\noindent
First, participants are generally unable to correctly differentiate between human-generated and AI-generated text, and there is no relationship between subject beliefs and actual LLM involvement. Second, despite being by and large uninformative, donors tend to condition their behaviour on these beliefs. These patterns are qualitatively present across the demographics of our subject pool, but primarily driven by Spanish-speaking female participants. For this subset, beliefs in AI-generated content substantially increase the share of people who act fully selfish, and strongly decrease absolute donations. This points to important cultural and gender effects in the evaluation of AI-generated charitable advertisement content. As female participants exhibit much stronger altruistic preferences, they may also react more elastically towards non-human involvement in charity work. \revision{This aligns with prior research demonstrating the significant role of gender in shaping perceptions of AI-generated content, particularly the heightened sensitivity of female participants \cite{zou2024pilot,zhang2023human}, and their concurrent stronger human preferences \cite{bellaiche2023humans}. Regarding potential cultural differences in the context of generative AI, the existing literature is scarce. Moreover, because we do not causally manipulate any cultural variables, restrict our analysis to a US-American sample, and do not observe specific cultural differences between our participants, our study does not provide any conclusions about the \textit{reasons} behind the difference between English and Spanish speaking donors. Instead, our results point to the existence of unobserved cultural endogeneity that may shape how people react to AI output. This relates, for example, to recent evidence that Western countries tend to be more critical and view generative AI as less aligned than Eastern countries \cite{globig2024perceived,arora2024pessimism}, or that European Americans are less likely to prefer AI with capabilities to influence, while exhibiting stronger preferences for control, than Chinese respondents \cite{ge2024culture}. In contrast, our study targets heterogeneity within the US-American society on a linguistic basis and therefore suggests that some cultural variables may influence AI perceptions on a more granular level. Finally, our findings contribute to a growing body of literature indicating that negative perceptions of AI-generated content are not necessarily merit-based, but often influenced by the belief that it was created by AI \cite{Harasta2024-jb,lim2024effect,shank2023ai}}.

\subsection{Implications}
As argued throughout this paper, our results relate to several interesting implications for different stakeholders. First, companies or industries that roll out LLM-based systems to different countries may want to consider \revision{evaluating the quality of their services post language add-on deployment, especially from a user behavioural perspective.}

Otherwise, exposure to mistakes in a particular native language may have broader consequences for the dissemination and adoption of their products. \revision{Especially those deploying multilingual LLM systems and benchmarking their performance in isolated, language-specific tests may overlook the cumulative impact of performance disparities across languages on user behaviour. For example, rolling out an LLM-based writing assistant without addressing cross-linguistic performance gaps could lead to disengagement among multilingual users, particularly in regions and domains where linguistic diversity and low(er)-resource languages are prevalent. This not only undermines the tool’s utility and subdues engagement metrics but also potentially exacerbates global inequities in AI adoption, as lower-resource language users tend to be disproportionately disadvantaged. From an efficiency standpoint, these patterns may inadvertently lead to substantial losses in productivity gains, as LLMs have been shown to be of particular use for relatively low-performing users \cite{vuculescu2024leveling,noy2023experimental,doshi2023generative}.} This highlights a potential downside of strategically deploying AI models ``early'' based only on benchmark data \cite{zhang2024dolares, jin2024better, conde2024opensourceconversationalllms}.

Beyond the suppliers, business consumers should also be aware that their employees who work in more than one language may be prone to under-utilization of these productive tools, and take respective countermeasures. We believe that these implications go beyond multilingual models, and extend towards a wide variety of AI and LLM models that are being used across various tasks. Hence, the integration of modern LLM systems into an organization's framework should systematically consider a model of human behaviour that considers strategic deviations from rationality and anticipates how exposure to heterogeneous stimuli affects decision-making.

Regarding responsible AI practices, it may be beneficial to explicitly consider user reactions towards performance shifts in the design of AI assistants. Beyond traditional questions about e.g., how to reliably communicate uncertainty or the sensitivity of certain contents, responsible design could think about alerting users to potential shifts in performance. While these may alert users who are otherwise unaware, they could also prevent negative second-order effects by endowing users with a reasonable interpretation of the nature of the performance shift. Furthermore, they can be used to clearly distinguish between different use cases, e.g., text generation between languages, and thereby potentially avoid detrimental generalization patterns. Practitioners themselves are likely to bear large parts of the costs induced by violations of choice independence, as their productivity suffers. Hence, being aware of such biases and communicating them could endow workers with the necessary tools to change their decision-making.

For developers, traditional software evaluation methods, such as unit testing, integration testing and system testing have been the fundamentals of system validation \cite{jamil2016software}. However, as AI systems become increasingly integrated with software, the focus of evaluation may need to shift. While the standard practice regarding AI evaluation has been to benchmark these ``black box" systems using extensive datasets across a range of scenarios, the practical utilization of AI-driven tools often hinges on user perceptions -- which can be heavily influenced by their rationality. The first stage of our study highlights the influence of such lapses in rational behaviour through a reduction in usage and interaction with the underlying system. By introducing frameworks from rational choice theory, particularly through the lens of choice independence, developers can gain valuable insights into system utilization. This approach not only helps improve feature usage but also fosters innovation by aligning system performance with user expectations.

For societies, there are implications within and between different countries. Within societies, there may be a divergence in productivity between different workers, depending on which languages they are being exposed to. For example, while native German workers in Germany have access to two high-resource languages, immigrants from some Eastern European countries could learn about the usefulness of LLMs through experiences in relatively lower-resource languages, with subsequent effects on everyday productivity. More generally, LLMs have been shown to provide large marginal benefits to relatively low-skilled workers. In so far as lower skills are correlated with certain marginalized languages, it may inhibit the anticipated reduction in inequality. Between societies, inequalities may also increase, as non-English countries in general, and smaller countries with less representation in particular, are endowed with lower quality LLMs that affect user beliefs beyond what would be rational.

\subsection{Caveats, Limitations, and Future Work}
\textbf{LLM and Tool Selection.} This study relies on a single LLM-augmented co-writing tool and an underlying LLM system. Both are not specifically optimized for persuasive writing, and results may differ for future tools that are more refined. In addition, while the pre-defined recipes are informed by the literature on charitable giving, there is still much to learn about what persuades people in the social preference domain, limiting what guidance we could offer participants. Consequently, recipes were not particularly popular, and it is plausible that LLMs with more refined features may provide more utility.

\noindent
\textbf{Writers.} Due to selection and availability constraints related to English-Spanish multilingual workers on Prolific, the number of writers in our first experiment was limited.
To make this study possible, we also had to define ``writer" more broadly as someone working in a writing-related profession. Future research can consider a larger and more specialized subject pool and control for writing expertise. 

\noindent
\textbf{Writing Task.} We commission short advertisements designed to elicit donations for an animal charity. This induces two limitations. One, due to the kind and size of the writing, we limit skill expression, which also constrains the influence of our AI assistant. Hence, under-utilization due to violations of choice independence may have substantially stronger effects in more sophisticated writing tasks. Two, we focus on persuasive writing in a social preference context. While this is a very important domain, it is also notably difficult to change peoples' social preferences. Future work may consider different applications, like sales, narrative text such as poems, or translations, for which we would expect a stronger influence of LLM-generated text on subsequent behaviour or judgments. It is also important to explore how multilingual LLMs shape user trust and reliance in increasingly popular agentic workflows that require complex planning and execution~\cite{he2025plan}.

\noindent
\textbf{Language Selection.} Due to availability constraints as described earlier, we compared two high-resource languages. Therefore, our results \revision{could} be interpreted as a lower baseline. It is plausible that writers who are being exposed to the LLM's performance in a true low-resource language are even more likely to generalize these experiences, resulting in larger behavioural changes. Hence, our study may well under-estimate the detrimental effect of between-language variation on more marginalized languages, and thereby the negative implications for inequality. \revision{On the other hand, more pronounced differences could increase users propensity to clearly differentiate between languages, and thus reduce cognitive interdependencies between AI performances across different tasks. These questions cannot be answered by our study.} Beyond that, behavioral patterns may also differ depending on the perceived similarity of the experienced languages. For example, people may be more likely to generalize errors from a Spanish LLM to an English LLM than to generalize errors from a Chinese LLM to an English LLM, as the latter two belong to different language families.

\noindent
\textbf{Prior Experience and Time Horizon.} We do not consider panel data. Therefore, we cannot draw any conclusions about how writers may adapt over time, whether a competitive market ``induces" more rational behaviour over time, or how prior experience with LLMs shapes the prevalence of choice independence violations. \revision{In particular, participants' reactions to heterogeneous stimuli across tasks may be mediated by their AI literacy levels, and the concurrent ability to contextualize, explain and maybe predict such differences. Importantly, knowledge about (multilingual) generative AI is not exogenous but can be affected through knowledge dissemination and policy. There may also be relevant differences across different societal or global groups.} These are all rich avenues for future research.

\noindent
\textbf{Cultural and Demographic Effects.} This study points to different reactions towards LLM-generated content between demographic groups. \revision{However, we are not able to distinguish the specific factors determining these differences. Future work could take a more targeted and controlled approach to tease-out specific causal factors behind different behavioral attitudes towards generative AI or their content across groups.} \revision{Moreover}, our work is limited to English and Spanish-speaking individuals who reside in the USA. Therefore, one natural extension is to expand the analysis towards people from different linguistic, cultural, and educational backgrounds. 

\noindent
\revision{\textbf{Cumulative vs. Sequential Effects and Temporal Impact.} Lastly, our study focuses on sequential second-order effects. However, the results may be influenced by participants' prior exposure to similar systems, particularly in Stage 1 experiments, where such pre-existing familiarity could shape observed behaviours.  Additionally, temporal effects, such as whether the observed usage behaviours persist beyond the study's duration, are beyond the scope of this research. Future work could explore these aspects to better understand the long-term implications of our findings.}
\section{Conclusions}
This study explores how multilingual LLMs affect user behaviour and advertisement persuasiveness in a co-writing task, with a focus on English and Spanish. Our findings reveal a clear ``rationality gap'', where prior exposure to a relatively lower-resource language (Spanish) diminishes subsequent reliance on an LLM writing assistant in a higher-resource language (English). We then evaluate the consequence of these patterns for the persuasiveness of generated ads via a charitable donation task. Here, peoples' donation behaviour appears largely unaffected by the specific advertisement, alleviating the potential negative consequences of under-utilization due to irrational behavioural adaptions. However, donations are strongly related to participants' beliefs about the source of the advertisement. Those who think that it was written by an AI are (1) significantly less likely to donate, and (2) donate less. Our results have strong implications for a number of important stakeholders, including companies deploying global multilingual AI assistants, the dissemination of LLMs across linguistically different parts of the world, marketing practitioners, and societal stakeholders concerned about inequality. Heterogeneity in AI performance across tasks can lead to substantial behavioural second-order effects that asymmetrically affect appropriate reliance and utilization.

\bibliographystyle{ACM-Reference-Format}
\bibliography{references}
\section{Appendix}
\label{Appendix}

\subsection{AI modifier prompts}
\label{subsec:ai_modifier_prompt}
\begin{itemize}
    \item \textbf{Positive Narrative Framing:} \\
    Prompt: Rewrite the sentence(s) to include a positive anecdotal story that highlights the benefits.
    
    \item \textbf{Cautionary Narrative Framing:} \\
    Prompt: Rewrite the sentence(s) to include a negative anecdotal story that highlights the consequences of not taking action.
    
    \item \textbf{Positive Narrative Framing with Statistics:} \\
    Prompt: Rewrite the sentence(s) to emphasize the positive outcomes using statistics.
    
    \item \textbf{Cautionary Narrative with Statistics:} \\
    Prompt: Rewrite the sentence(s) to highlight the negative consequences using statistics.
    
    \item \textbf{Emphasis on Immediate Outcome:} \\
    Prompt: Rewrite the sentence(s) to emphasize the immediate impact of the donation.
    
    \item \textbf{Emphasis on Long-Term Outcome:} \\
    Prompt: Rewrite the sentence(s) to emphasize the long-term impact of the donation.
\end{itemize}

\subsection{Average Word Count and Time Spent Outside the Writing Environment}
\begin{figure}[H]
    \centering
    \includegraphics[width=0.45\linewidth]{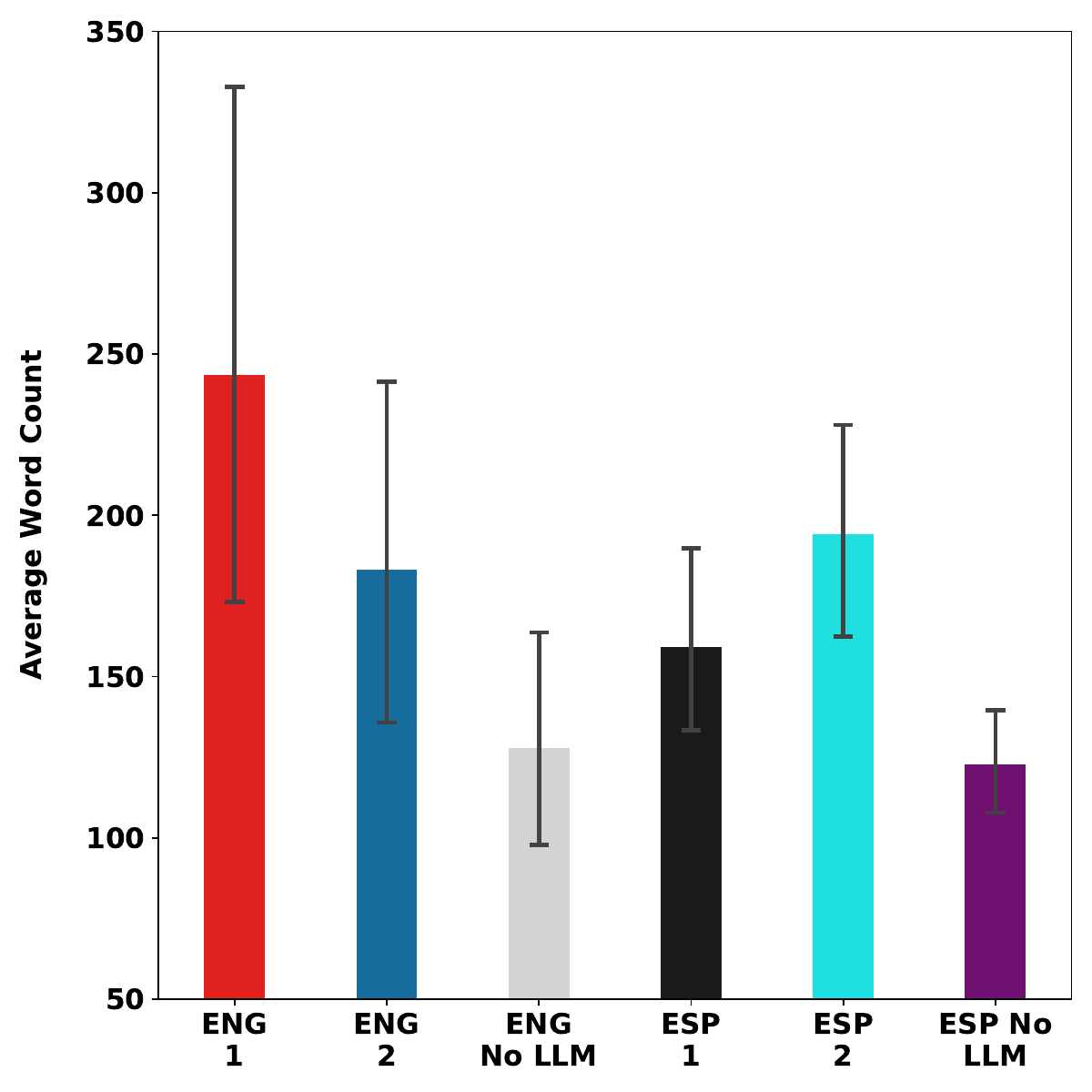}
    \includegraphics[width=0.45\linewidth]{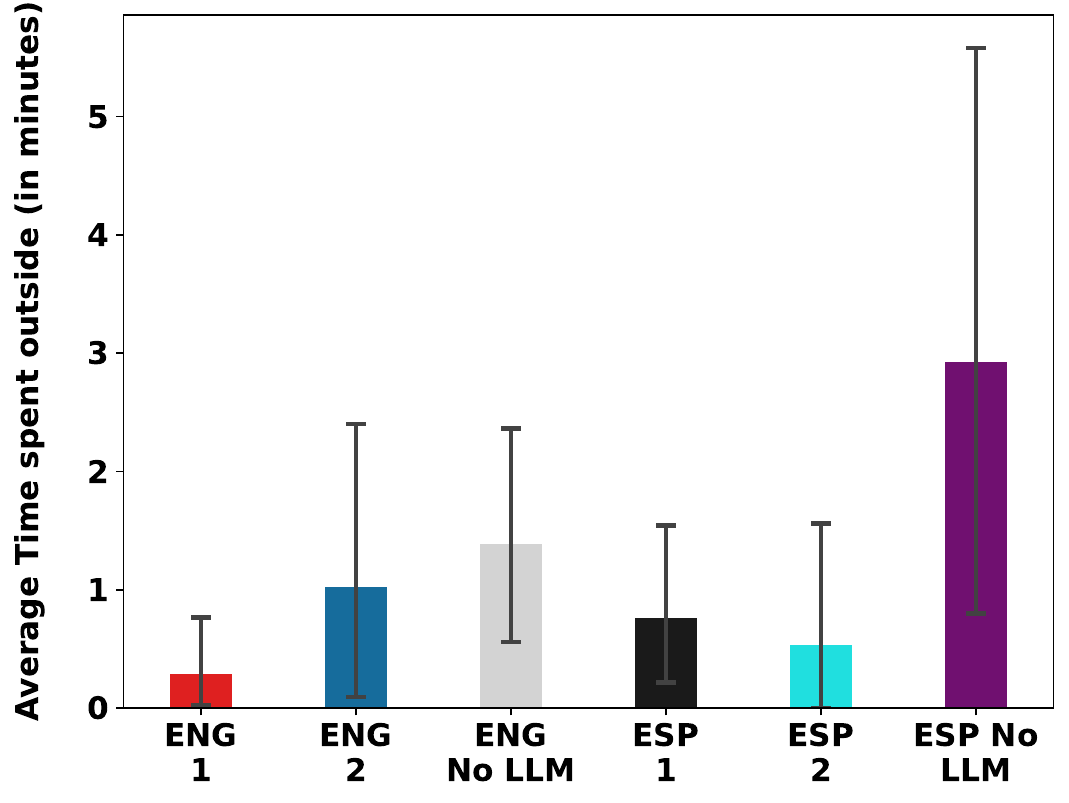}
    
     \Description{A bar graph showing average word count across different tasks, with error bars representing confidence interval.}
    \caption{\revision{\textbf{Left:} Average word count across treatments, indicating variations in output length depending on treatment and condition. \textbf{Right:} Average time spent outside the platform across treatments. Error bars represent 95\% confidence intervals.}}
    \label{fig:stage1_word_count}
\end{figure}

\subsection{Stage 1 - Writing Task Metrics: Stated utility for all features}

\begin{figure}[H]
    \centering
    \includegraphics[width=0.45\linewidth]{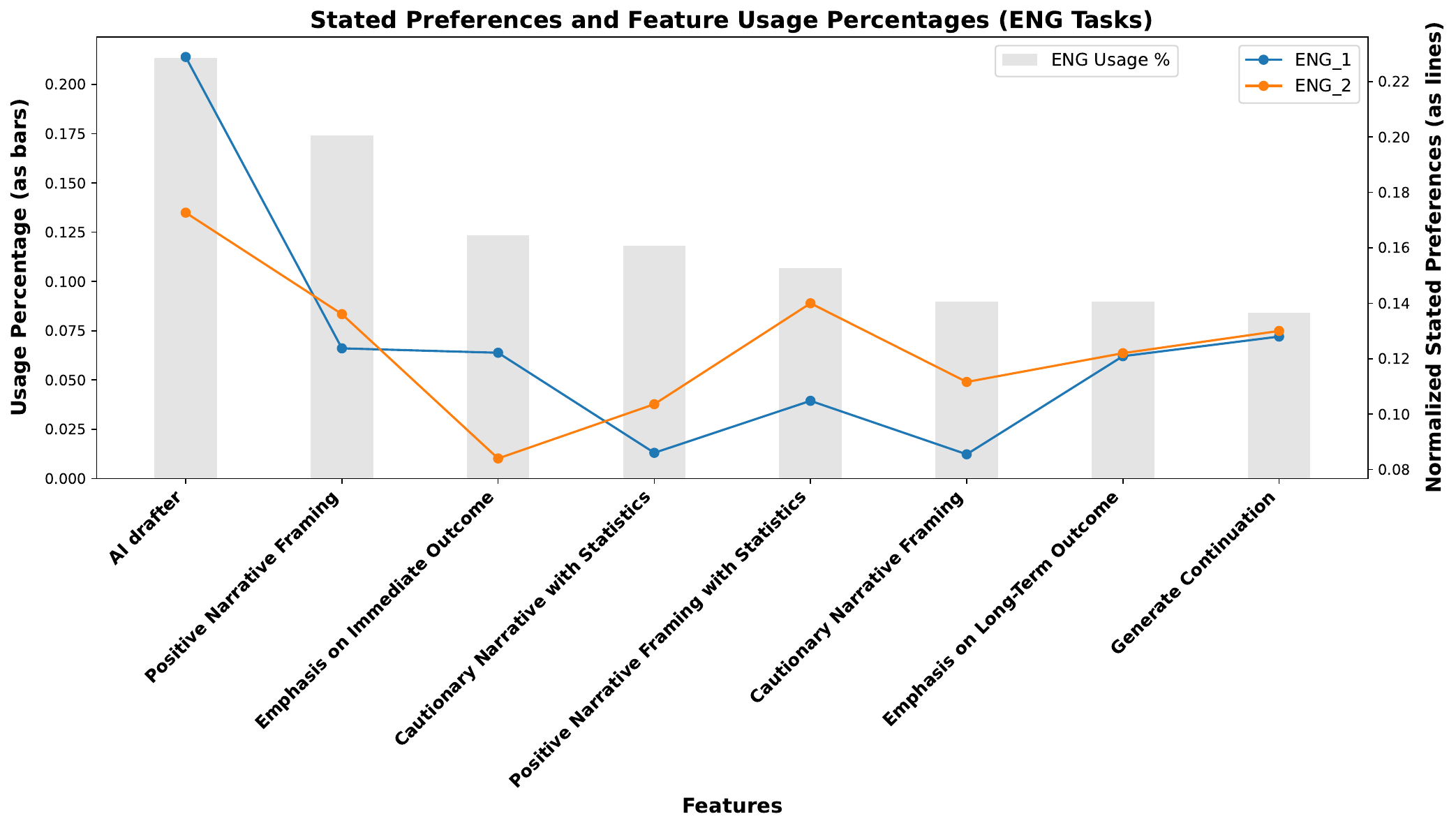}
    \includegraphics[width=0.45\linewidth]{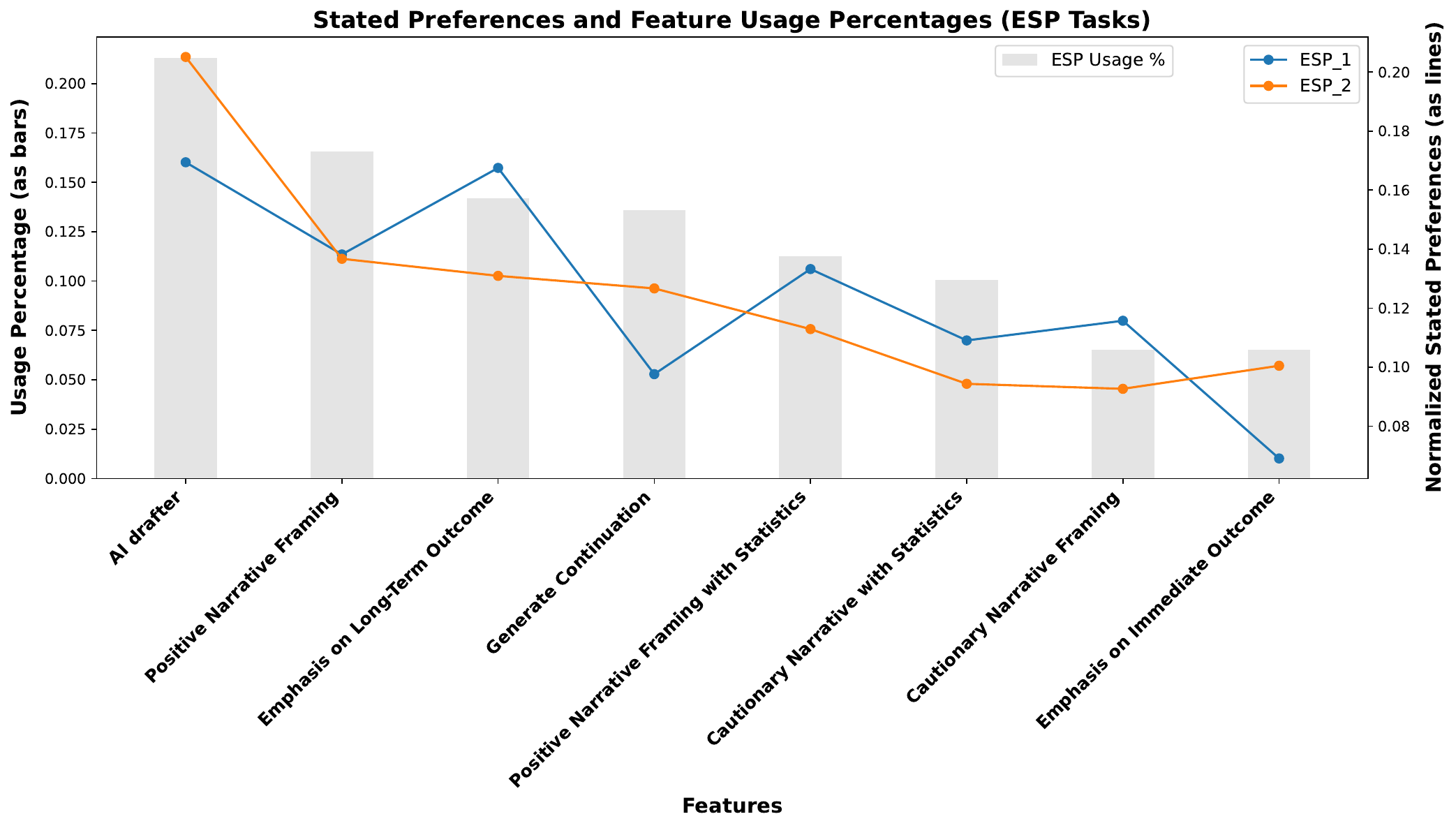}
    \caption{\revision{Differences in stated utility and feature usage percentages across treatment groups for English (Left) and Spanish (Right) tasks. Bars represent feature usage percentages, while lines indicate normalized stated preferences as reported in questionnaire responses. This visualization highlights discrepancies between stated utility and actual usage across language conditions.}}
    \label{fig:stated_utility}
\end{figure}

\subsection{Stage 1 - Writing Task Metrics: Revealed utility for all features}
\begin{figure}[H]
    \centering
    \includegraphics[width=0.45\linewidth]{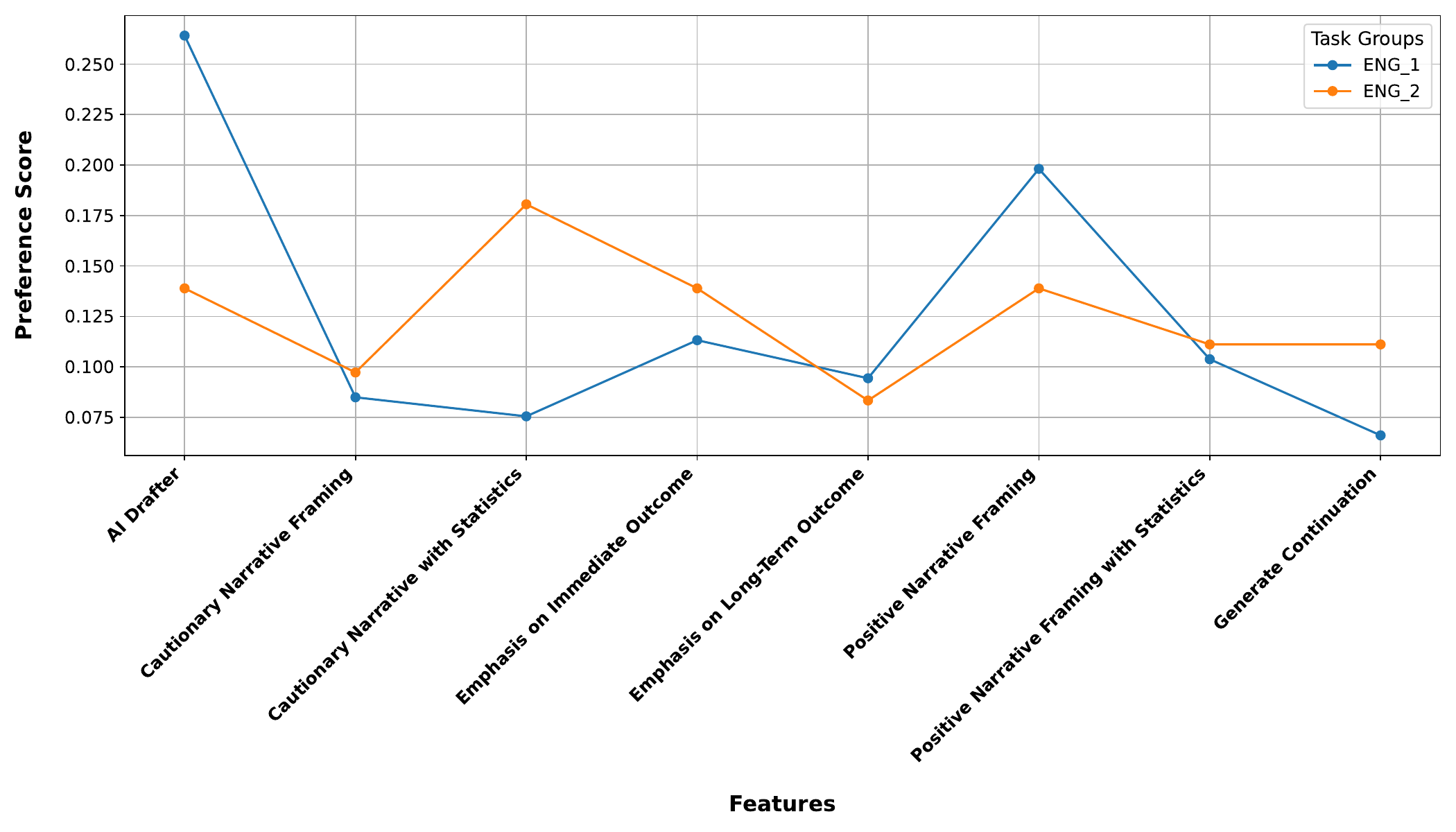}
    \includegraphics[width=0.45\linewidth]{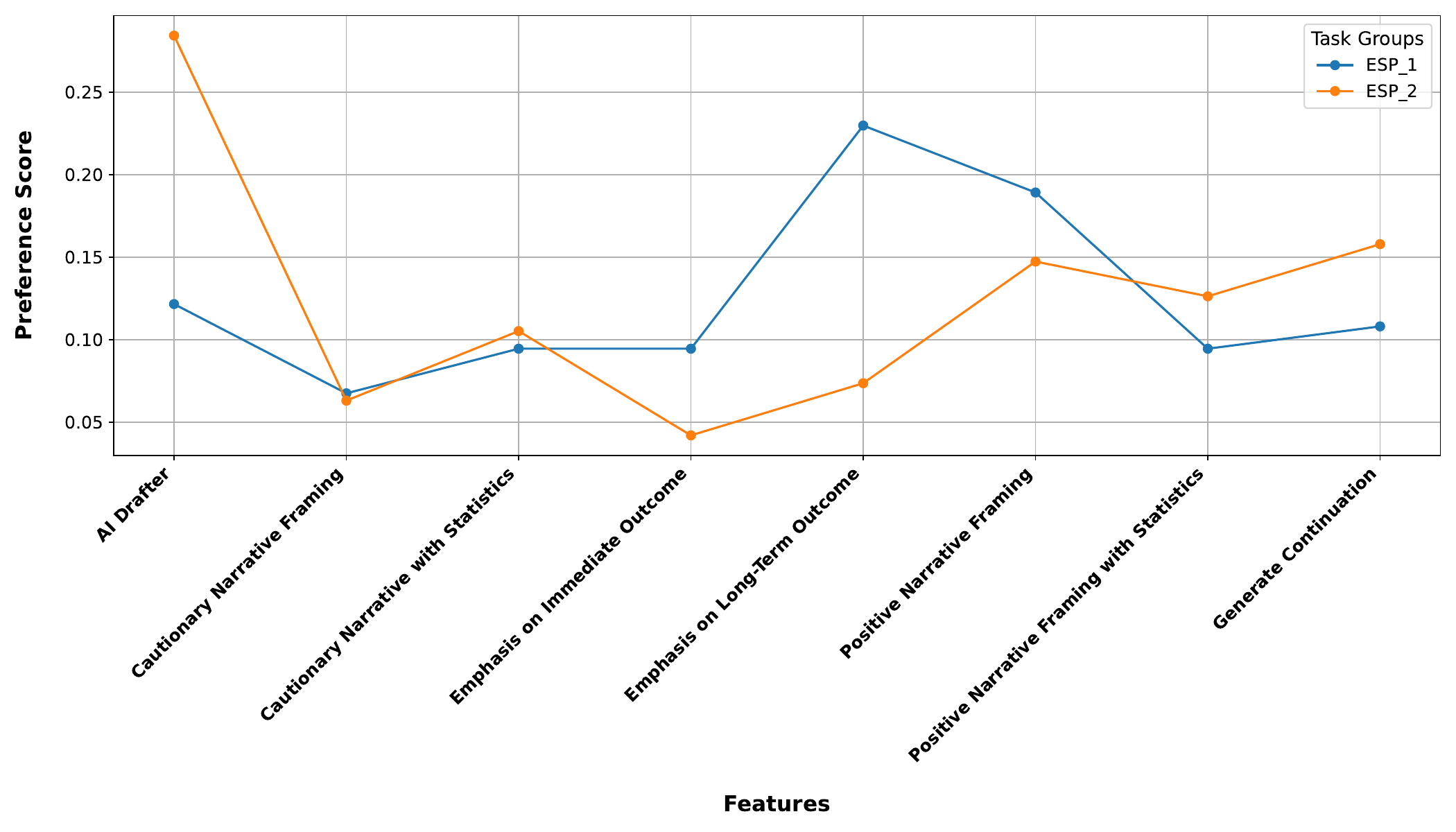}
    
    \caption{\revision{Differences in actual utility across treatment groups for English (Left) and Spanish (Right) tasks, represented as preference scores derived from usage data. The figure illustrates how features were prioritized by participants in different task conditions (ENG\_1, ENG\_2 for English; ESP\_1, ESP\_2 for Spanish), highlighting variations in feature utility across languages and conditions.}}
    \label{fig:revealed_utility_button_clicks_eng}
\end{figure}

\subsection{Stage 2: Donation Task Metrics}

\begin{figure}[H]
    \centering
    \includegraphics[width=0.3\linewidth]{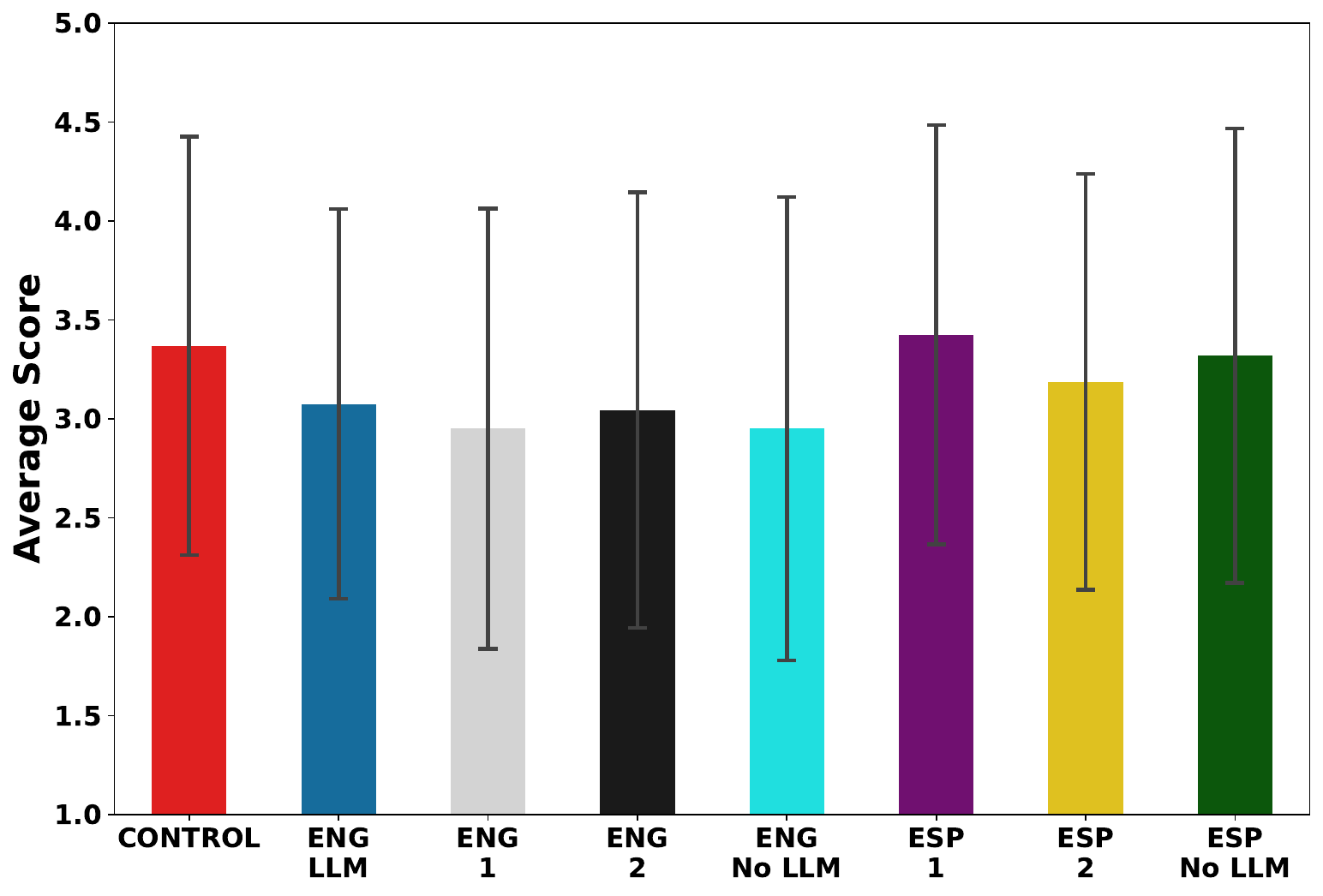}
    \includegraphics[width=0.3\linewidth]{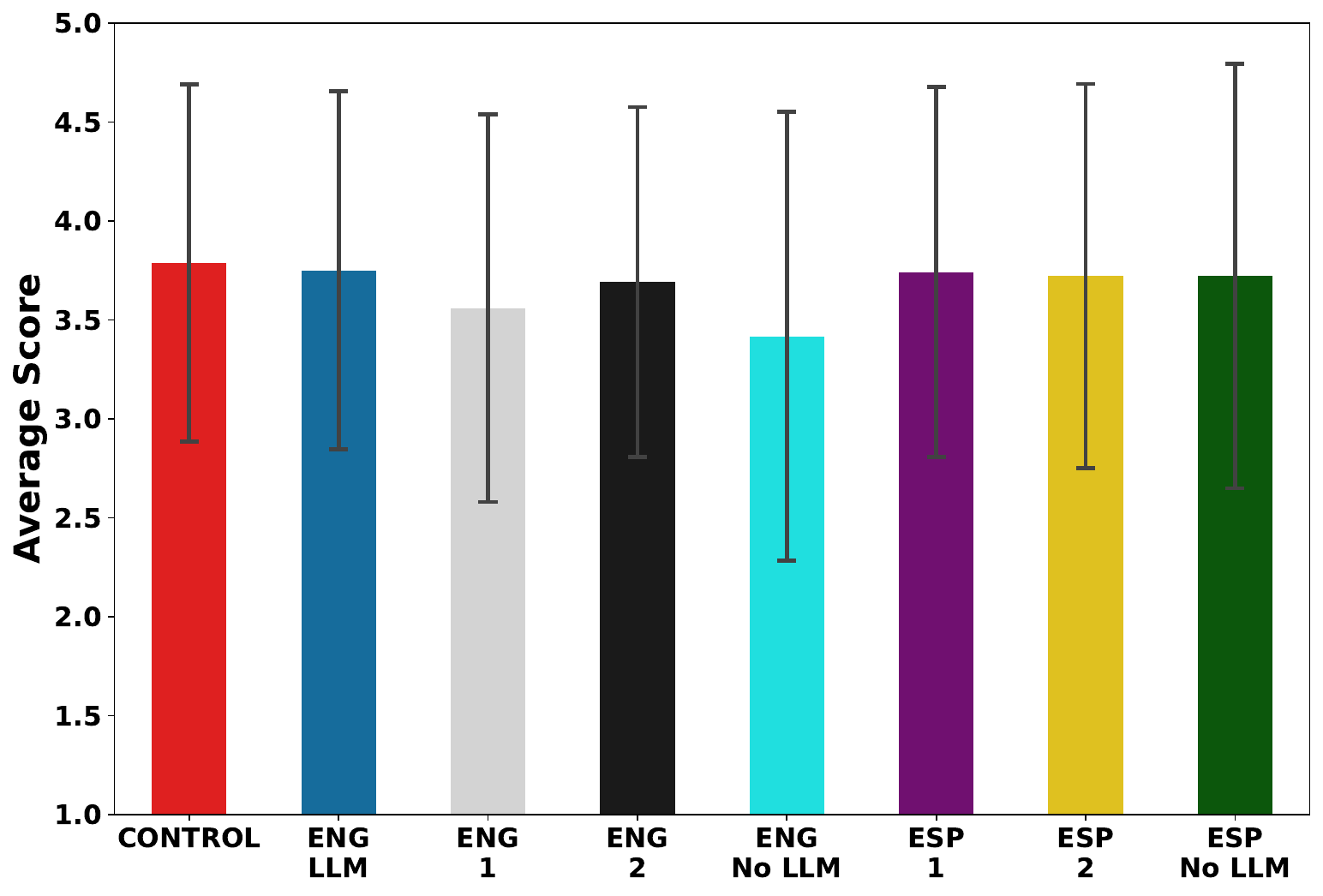}
    \includegraphics[width=0.3\linewidth]{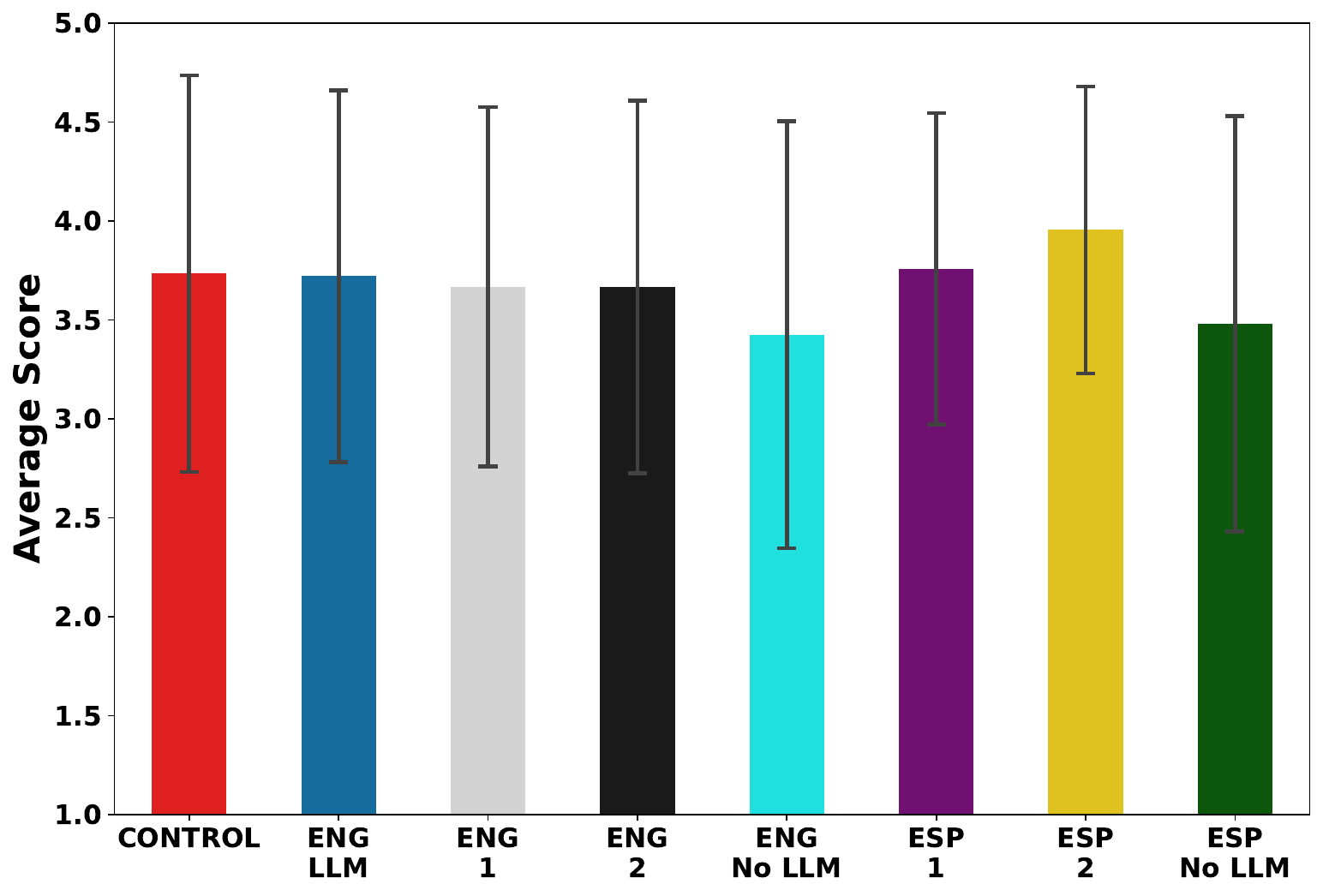}

     \Description{A graph showing the distribution of perceptions of Behavioral Intention for texts across different experimental conditions.}
    \caption{\revision{Average scores for \textbf{Behavioral Intention} (left), \textbf{Emotional Appeal} (middle), and \textbf{Information Awareness} (right) across various experimental conditions. Error bars represent 95\% confidence intervals.}}

    \label{fig:stage2_bi}
\end{figure}

\subsection{Stage 2: Donation Behaviour by Demographic Factors}

\begin{figure}[H]
    \centering
    \includegraphics[width=0.45\linewidth]{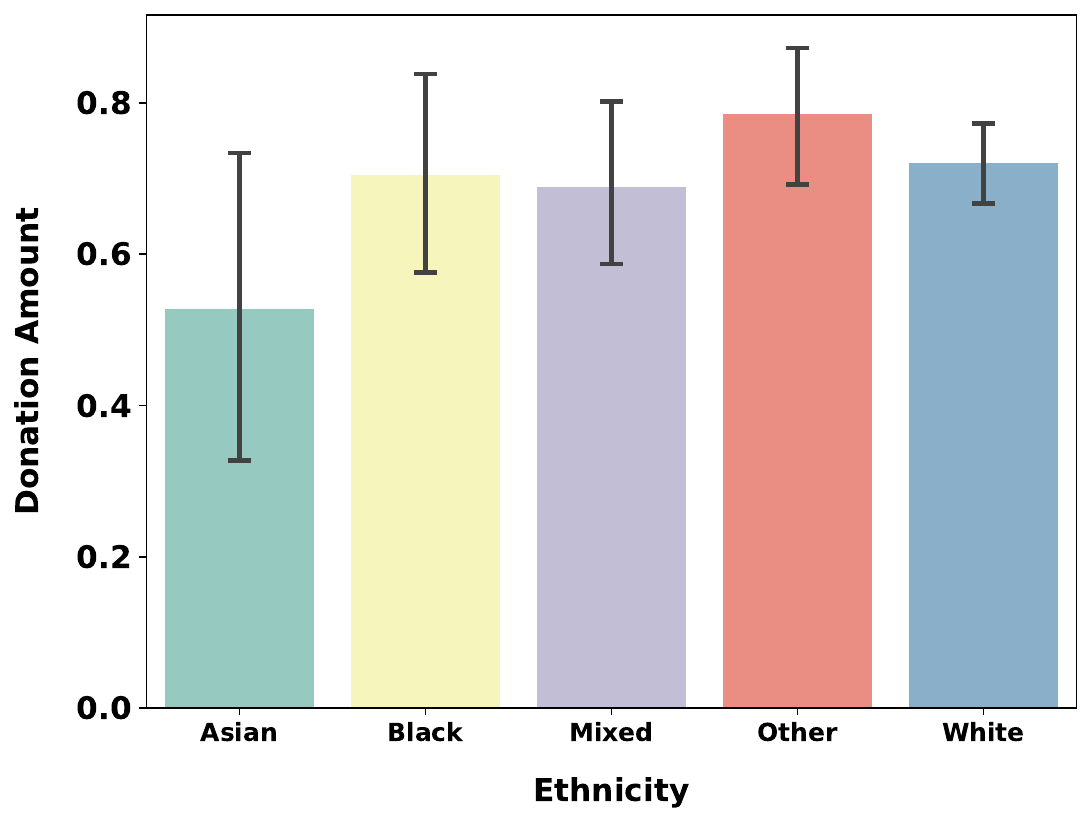}
    \includegraphics[width=0.45\linewidth]{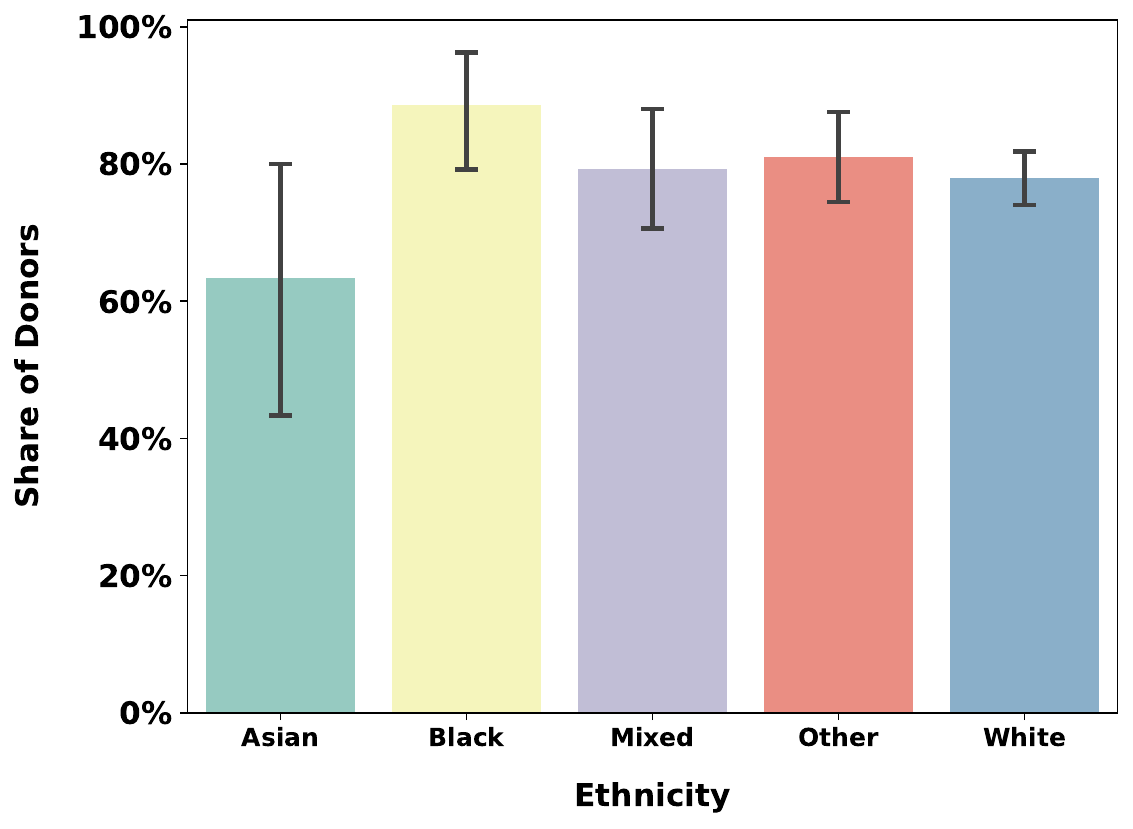}
     \Description{A graph showing the distribution of donation behaviour by Ethnicity.}
    \caption{\revision{\textbf{Left:} Average donation amount across different ethnic groups. \textbf{Right:} Share of donors (percentage of participants who donated) by ethnicity, highlighting variations in donation behavior and likelihood among demographic groups. Error bars representing 95\% confidence intervals.}}
    \label{fig:stage2_ethnicity}
\end{figure}

\begin{figure}[H]
    \centering
    \includegraphics[width=0.45\linewidth]{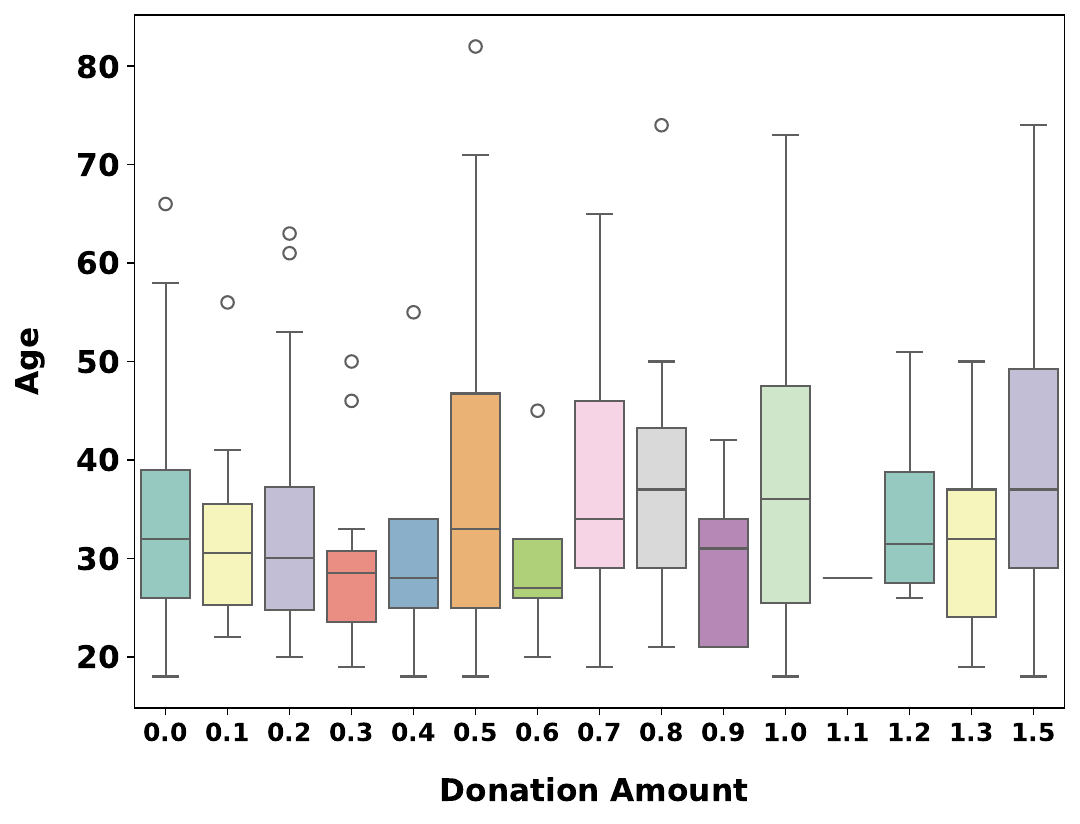}
    \includegraphics[width=0.45\linewidth]{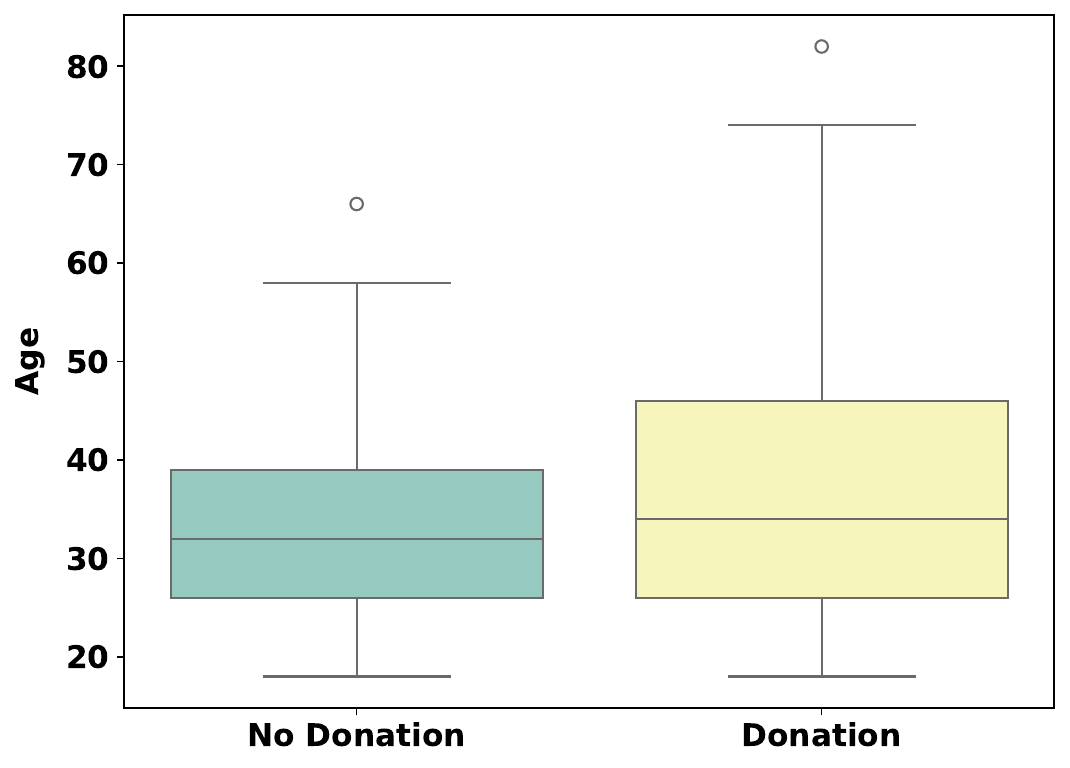}
     \Description{A graph showing the distribution of donation behaviour by Age.}
    \caption{\revision{\textbf{Left:} Distribution of donation amounts across age groups, showing the variability in donation behavior as a function of age. \textbf{Right:} Comparison of age distributions between donors and non-donors, highlighting differences in median age and interquartile ranges for both groups}}
    \label{fig:stage2_age}
\end{figure}

\begin{table}[t]
\centering
\caption{\revision{Results from an Ordinary Least Squares (OLS) regression model examining the effects of demographic and contextual factors on \textbf{donation amount}.}}
\label{tab:demograph_donation_amount}
\begin{tabular}{lrrrr}
\toprule
\textbf{} & \textbf{Coefficient} & \textbf{SE} & \textbf{t-value} & \textbf{95\% CI} \\
\midrule
Age (normalized) & 0.096*** & 0.021 & 4.481 & [0.054, 0.138] \\
Sex (Male) & -0.229*** & 0.041 & -5.599 & [-0.309, -0.149] \\
Language (Spanish) & 0.003 & 0.051 & 0.051 & [-0.098, 0.103] \\
\addlinespace[3pt]
Ethnicity\textsuperscript{a} &&&&\\
\quad Black & 0.177 & 0.133 & 1.324 & [-0.085, 0.438] \\
\quad Mixed & 0.138 & 0.122 & 1.131 & [-0.102, 0.378] \\
\quad Other & 0.273* & 0.123 & 2.228 & [0.032, 0.513] \\
\quad White & 0.155 & 0.111 & 1.398 & [-0.063, 0.373] \\
\addlinespace[3pt]
Region of birth\textsuperscript{b} &&&&\\
\quad Asia & 0.320 & 0.242 & 1.323 & [-0.155, 0.794] \\
\quad Europe & 0.012 & 0.216 & 0.056 & [-0.413, 0.437] \\
\quad North America & 0.110 & 0.161 & 0.687 & [-0.205, 0.425] \\
\quad South America & 0.130 & 0.180 & 0.723 & [-0.223, 0.484] \\
\addlinespace[3pt]
Belief Ad Source (Human) & 0.088* & 0.040 & 2.186 & [0.009, 0.167] \\
Intercept & 0.492* & 0.195 & 2.520 & [0.109, 0.875] \\
\bottomrule
\multicolumn{5}{l}{\textit{Note.} SE = Standard Error; CI = Confidence Interval.} \\
\multicolumn{5}{l}{\textsuperscript{a,b}Reference categories are "Asian" and "Africa" respectively. *p < .05. **p < .01. ***p < .001.} \\
\end{tabular}
\vspace{1em}

\caption{\revision{Results from a logistic regression model assessing the impact of demographic and contextual variables on the \textbf{likelihood of a person donating}.}}
\label{tab:demograph_is_donor}
\begin{tabular}{lrrrr}
\toprule
\textbf{} & \textbf{Coefficient} & \textbf{SE} & \textbf{z-value} & \textbf{95\% CI} \\
\midrule
Age (normalized) & 0.432*** & 0.120 & 3.589 & [0.196, 0.667] \\
Sex (Male) & -0.820*** & 0.197 & -4.151 & [-1.207, -0.433] \\
Language (Spanish) & 0.242 & 0.249 & 0.971 & [-0.247, 0.731] \\
\addlinespace[3pt]
Ethnicity\textsuperscript{a} &&&&\\
\quad Black & 1.112 & 0.641 & 1.734 & [-0.145, 2.369] \\
\quad Mixed & 0.388 & 0.520 & 0.746 & [-0.631, 1.407] \\
\quad Other & 0.602 & 0.527 & 1.142 & [-0.430, 1.634] \\
\quad White & 0.323 & 0.456 & 0.708 & [-0.571, 1.217] \\
\addlinespace[3pt]
Region of birth\textsuperscript{b} &&&&\\
\quad Asia & -0.527 & 1.361 & -0.387 & [-3.195, 2.140] \\
\quad Europe & -1.568 & 1.271 & -1.234 & [-4.059, 0.923] \\
\quad North America & -0.910 & 1.118 & -0.814 & [-3.101, 1.280] \\
\quad South America & -0.044 & 1.234 & -0.036 & [-2.462, 2.374] \\
\addlinespace[3pt]
Belief Ad Source (Human) & 0.562** & 0.203 & 2.766 & [0.164, 0.960] \\
Intercept & 1.862 & 1.208 & 1.542 & [-0.505, 4.229] \\
\bottomrule
\multicolumn{5}{l}{\textit{Note.} N = 692. SE = Standard Error; CI = Confidence Interval.} \\
\multicolumn{5}{l}{\textsuperscript{a,b}Reference categories are "Asian" and "Africa" respectively. *p < .05. **p < .01. ***p < .001.} \\
\end{tabular}
\end{table}

\subsection{Language Benchmarking Results}

\begin{table}[H]
\centering
\caption{\revision{Language Performance metrics across benchmarks for English and Spanish for LLama 3.1-8B model. Benchmarks include Multi-IF \cite{he2024multi} Accuracy, Paraphrase Adversaries from Word Scrambling (PAWS-X) \cite{pawsx2019emnlp}, and Multilingual Persuasion Detection datasets \cite{Pyhnen2022MultilingualPD} (ROC/PR AUC). [We did not consider accuracy as the labels in this dataset are highly skewed]. The prompt constructions to run these benchmarks were inspired by \citet{Ahuja2023-zu}}}
\resizebox{\textwidth}{!}{%
\begin{tabular}{|l|c|c|c|c|}
\hline
\textbf{Language} & \textbf{Multi-IF  Accuracy (Turn 1/2/3)} & \textbf{Paraphrasing (PAWS-X)} & \textbf{ROC-AUC (Persuasive)} & \textbf{PR-AUC (Persuasive)} \\ \hline
English & 0.75 / 0.66 / 0.57 & 51\% & 0.38 & 0.20 \\ \hline
Spanish & 0.43 / 0.44 / 0.42 & 54\% & 0.29 & 0.17 \\ \hline
\end{tabular}%
}
\label{tab:benchmark_results}
\end{table}

\subsection{Stage 2: Donation Task Questionnaire}

The Table~\ref{tab:questionnaire} is available in the next page.

\begin{table}[httb]
\centering
\caption{Summary of donation task questionnaire}
\label{tab:questionnaire}
\scalebox{.75}{
\begin{tabular}{|p{5cm}|p{7cm}|p{3cm}|}
\hline
\textbf{Category} & \textbf{Question/Description} & \textbf{Response Scale} \\ 
\hline
\textbf{Emotional Appeal} & 
\begin{itemize}[nosep]
    \item The advertisement made me feel emotionally connected to the cause.
    \item I felt a strong sense of empathy for the cause featured in the advertisement.
    \item The ad evoked feelings of compassion and a desire to help.
\end{itemize} &  5-point Likert \\ 
\hline

\textbf{Information and Awareness} & 
\begin{itemize}[nosep]
    \item The advertisement provided useful information about the organization.
    \item I feel more informed about the organization's work after seeing the ad.
    \item The ad increased my understanding of the impact the organization has.
\end{itemize} &  5-point Likert \\ 
\hline

\textbf{Behavioral Intention} & 
\begin{itemize}[nosep]
    \item How likely are you to share this advertisement or talk about the cause with others?
    \item How likely are you to seek more information about the organization?
\end{itemize} &  5-point Likert \\ 
\hline

\textbf{Attention Check} & 
Which charity have you donated to in this task?
\begin{itemize}[nosep]
    \item WWF
    \item Red Cross
    \item UNICEF
    \item Immediate impact
    \item Doctors Without Borders
\end{itemize} & Single Choice \\ 
\hline

\textbf{Positive Aspects} & 
What aspects did you like about the charity ad and why? & Open-ended \\ 
\hline

\textbf{Negative Aspects} & 
What aspects did you NOT like about the charity ad and why? & Open-ended \\ 
\hline

\textbf{Perceived Source of the Ad} & 
Who do you think wrote the ad? (AI or Human) & Single choice \\ 
\hline

\textbf{Reason of Selecting Source of the Ad} & 
Why did you select (AI or Human)?  & Open-ended \\ 
\hline

\textbf{Key Features of the Ad (recipe usage)} & 
What caught your attention the most in the ad? (Options are shortened for brevity)
\begin{itemize}[nosep]
    \item Positive story
    \item Warning story
    \item Statistics
    \item Immediate impact
    \item Long-term impact
    \item None
\end{itemize}
& Multiple choice \\ 
\hline

\end{tabular}}
\end{table}

\end{document}